\documentclass[10pt,twocolumn,letterpaper]{article}
\usepackage{cvpr}              
\usepackage{color,array}
\usepackage{colortbl}
\usepackage{cite}
\usepackage{graphicx}
\usepackage{amsmath}
\usepackage{amsfonts}
\usepackage{amssymb}
\usepackage{bbm}
\usepackage{algorithm}
\usepackage{algorithmic}
\usepackage{xcolor}
\usepackage{amssymb}
\usepackage{multirow}
\usepackage{array}
\usepackage{stfloats}
\usepackage{verbatim}
\usepackage{graphics}
\usepackage{booktabs}
\usepackage{mathrsfs}
\usepackage[precent]{overpic}
\usepackage{booktabs}

\usepackage{indentfirst}
\usepackage{enumitem}
\definecolor{commentcolor}{HTML}{d62c2c} 
\definecolor{forestgreen}{HTML}{009B55} 

\definecolor{shallowgray}{HTML}{738385}

\definecolor{cvprblue}{rgb}{0.21,0.49,0.74}
\definecolor{Gray}{gray}{0.85}

\usepackage[pagebackref,breaklinks,colorlinks,citecolor=cvprblue]{hyperref}

\def\paperID{6899} 
\def\confName{CVPR}
\def\confYear{2024}

\title{Rethinking Multi-view Representation Learning via Distilled Disentangling}

\author{
Guanzhou Ke$^{1,}$\thanks{This work was done while interning at the Institute of Automation, Chinese Academy of Sciences.} , Bo Wang$^{2,}$\thanks{Corresponding authors.} , Xiaoli Wang$^3$, Shengfeng He$^{4,}$\textsuperscript{\textdagger}\\
$^1$Beijing Jiaotong University,\\ $^2$State Key Laboratory of Multimodal Artificial Intelligence Systems, \\
Institute of Automation, Chinese Academy of Sciences, \\
$^3$Nanjing University of Science and Technology, $^4$Singapore Management University \\
{\tt\small guanzhouk@gmail.com, wangbo@ia.ac.cn, xiaoliwang@njust.edu.cn, shengfenghe@smu.edu.sg}
}

\begin{document}
\maketitle

\begin{abstract}
Multi-view representation learning aims to derive robust representations that are both view-consistent and view-specific from diverse data sources. This paper presents an in-depth analysis of existing approaches in this domain, highlighting a commonly overlooked aspect: the redundancy between view-consistent and view-specific representations. To this end, we propose an innovative framework for multi-view representation learning, which incorporates a technique we term `distilled disentangling'.
Our method introduces the concept of \textit{masked cross-view prediction}, enabling the extraction of compact, high-quality view-consistent representations from various sources without incurring extra computational overhead. Additionally, we develop a distilled disentangling module that efficiently filters out consistency-related information from multi-view representations, resulting in purer view-specific representations.
This approach significantly reduces redundancy between view-consistent and view-specific representations, enhancing the overall efficiency of the learning process. Our empirical evaluations reveal that higher mask ratios substantially improve the quality of view-consistent representations. Moreover, we find that reducing the dimensionality of view-consistent representations relative to that of view-specific representations further refines the quality of the combined representations. {Our code is accessible at: \url{https://github.com/Guanzhou-Ke/MRDD}}.
\end{abstract}


\section{Introduction}
\label{sec:introduction}
Multi-view representation learning (MvRL) \cite{wang2015deep} forms the cornerstone of various multi-view applications, such as video understanding \cite{jin2016describing, chen2022mm}, 3D rendering \cite{zhou2020end-3d}, and cross-modal retrieval~\cite{rasiwasia2010new}. In the MvRL context, ``views'' commonly refer to distinct angles from which objects are captured by cameras or data descriptors, like the histogram of oriented gradients (HOG) \cite{dalal2005histograms} and the scale-invariant feature transform (SIFT) \cite{lowe1999object}. The success of multi-view applications relies on effectively leveraging shared information (\textit{consistency}) among views and distinctive information (\textit{specificity}) within each view. However, learning high-quality view-consistent and view-specific representations from multiple sources poses an open challenge.


\begin{figure}[t]
\setlength{\abovecaptionskip}{0cm} 
\setlength{\belowcaptionskip}{-0.2cm}
  \centering
   \begin{overpic}[scale=.48]{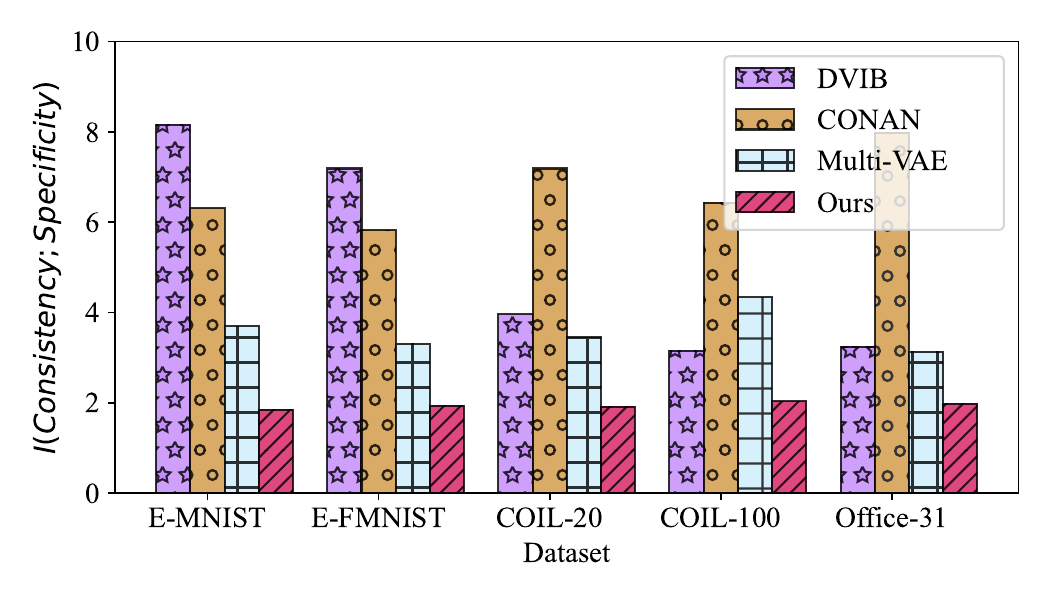}
    \put(90, 41){\rotatebox{0}{\scriptsize  ~\cite{xu2021multi}}}
    \put(84.5, 48.7){\rotatebox{0}{\scriptsize  ~\cite{bao2021disentangled}}}
    \put(87.5, 44.9){\rotatebox{0}{\scriptsize  ~\cite{ke2021conan}}}
   \end{overpic}\vspace{-2mm}
   \caption{Existing multi-view representation learning methods show high inter-view correlations. We estimate the mutual information of multi-view consistency and specificity of three baseline MvRL models DVIB~\cite{bao2021disentangled}, CONAN~\cite{ke2021conan}, Multi-VAE~\cite{xu2021multi}, and our method using MINE~\cite{belghazi2018mutual} on the same settings across five datasets.}
   \label{fig:observation}
\vspace{-10pt}
\end{figure}


Optimal view-consistent and view-specific representations should exhibit both robust representational abilities and minimal redundancy. Reducing redundancy between these two aspects not only improves the quality of combined representations but also decreases the computational burden for subsequent tasks. However, prevailing state-of-the-art methods~\cite{xu2022multi, federici2020learning, ke2023disentangling, bao2021disentangled, xu2021multi} often neglect the critical aspect of minimizing redundancy between consistency and specificity. As demonstrated in Fig.~\ref{fig:observation}, there is a notable correlation among representations derived from existing end-to-end approaches\footnote{More details are included in the supplementary materials.}. This leads us to pose an important question: \textit{What factors contribute to the dependency within multi-view representations?} In this paper, we address this question from a disentanglement perspective:

\textbf{\textit{(i) In the unsupervised setting, the joint learning (or end-to-end) paradigm presents significant challenges.}} Within the scope of disentangling representations, most end-to-end MvRL methods aim to extract view-consistent representations by maximizing their mutual information lower bound across views, while simultaneously minimizing the mutual information upper bound between views to derive view-specific representations. This approach essentially forms a Min-Max game, posing a risk for models to settle on suboptimal solutions in the absence of supplementary information. Recent advancements have sought to refine the model's proficiency in learning high-quality representations by incorporating auxiliary constraints into the joint loss function, such as adversarial~\cite{zhou2020end} and contrastive constraints~\cite{ke2021conan, trosten2021reconsidering, wang2024knowledge, ke2023clustering}. Although these approaches mitigate some limitations of joint representation learning, they often overlook a critical factor: the model's initial inability to differentiate between view-consistent and view-specific information, leading to the accumulation of redundancy.

We propose that if a model could effectively strip away information linked to pre-existing knowledge from multi-view representations, the remaining data would be devoid of such knowledge. We term this process `distilled disentangling.' In our methodology, view-consistent representations are treated as prior knowledge, under the rationale that such consistency represents information common to all views and remains unchanged irrespective of the view. By identifying and excluding the view-consistent information, the model can more accurately isolate view-specific representations for each view.

\textbf{\textit{(ii) Disparity in Information Density Between Consistency and Specificity.}} Extracting view-specific information typically involves processing individual views, whereas view-consistent information necessitates the integration of data from all views. Crafting a unified representation from multiple sources is complex and can lead to a significant escalation in computational resources as the number of views increases. To counter this challenge, some approaches employ multiple lightweight view-specific autoencoders to generate multi-view latent representations~\cite{xu2021multi, xu2022multi}, followed by an exploration of consistency using these synthesized representations. However, each latent representation is heavily laden with view-specific information, posing a challenge in mitigating this interference.

Our solution introduces the concept of \textit{masked cross-view prediction (MCP)}, which facilitates the learning of multi-view consistency using a single consistent encoder, without the need for additional computational resources. This is achieved by selectively masking parts of the content and prompting the encoder to predict the masked content by synthesizing visible portions from multi-view data. The advantages of MCP are manifold: 1) it efficiently processes all view data without escalating computational demands; 2) the randomness of masking aids in minimizing the impact of view-specific information; and 3) it strengthens the resilience of view-consistent representations.

Building on this analysis, we adeptly tackle the identified challenges. Leveraging the Masked Cross-View Prediction (MCP) strategy, we employ a single consistent encoder to process all views, yielding high-quality view-consistent representations. Initially, all unmasked blocks are fed into the consistent encoder simultaneously to derive these representations. This step is followed by utilizing multiple decoders to predict the masked content in their respective views, using the same view-consistent representation. The primary benefit of this method is its efficiency in extracting concise consistent representations while concurrently minimizing computational demands. Our experiments demonstrate that, even with a high mask ratio, for instance, 80\%, the MCP strategy enables the model to learn superior representations compared to the no-MCP baseline.

To address the issue (i), we freeze the consistent encoder post-training, maintaining the invariance of view-consistent representations during the distillation process for view-specific representations. Additionally, we design a disentangling module that minimizes the upper bound of mutual information between the consistent and view-specific representations, thereby extracting refined view-specific representations. To prevent trivial solutions in the disentangling module, we concatenate the consistent and view-specific representations, employing view-specific decoders to reconstruct the original data. These integrated approaches culminate in our novel multi-view representation learning method, which we term Multi-view Representation learning via Distilled Disentangling (MRDD). The key contributions of this paper are outlined as follows:
\begin{itemize}[leftmargin=0.3cm]
\item We illuminate fundamental challenges in multi-view representation learning through a disentanglement lens, revealing how these limitations impede the effectiveness of existing models.
\item We introduce a multi-view representation learning framework centered on distilled disentangling, which offers a fresh perspective on crafting low-redundancy view-consistent and view-specific representations. Our extensive experimental analysis confirms the superiority of our approach over current state-of-the-art methods.
\item Our experimental findings highlight two key insights: i) a high masked ratio (e.g., 80\%) significantly enhances the quality of consistent representations; ii) reducing the dimensionality of consistent representations relative to specific representations markedly boosts the quality of their combined representations. We believe that these discoveries will inspire further research in the field of MvRL.
\end{itemize}

\section{Related Work}
\label{sec:related-work}

\textbf{Multi-view Representation Learning.}
The goal of MvRL is to extract both shared and view-specific information from multiple data sources, integrating them into a cohesive representation that is advantageous for predictive tasks~\cite{LiYZ19, sun2013multi, chao2016consensus, ke2022efficient}. Existing approaches in this field generally fall into three categories: statistic-based~\cite{rasiwasia2010new, dalal2005histograms, liu2013multi, wang2018multiview, liu2013multi, wang2018multiview}, deep learning-based~\cite{andrew2013deep, wang2015deep, zhang2019ae2, ke2021conan, trosten2021reconsidering, xu2023progressive}, and hybrid methods~\cite{huang2021deep, sun2019self, wang2021dense}. More discussions are given in the supplementary materials.


Our approach is categorized under deep learning-based methods. We distinguish our work by utilizing deep learning's capacity to handle large datasets effectively. Moreover, we address the interpretability challenges in representations by incorporating disentanglement techniques

\begin{figure*}[t]
\setlength{\abovecaptionskip}{0cm} 
\setlength{\belowcaptionskip}{-0.2cm}
  \centering
   \includegraphics[width=1\linewidth]{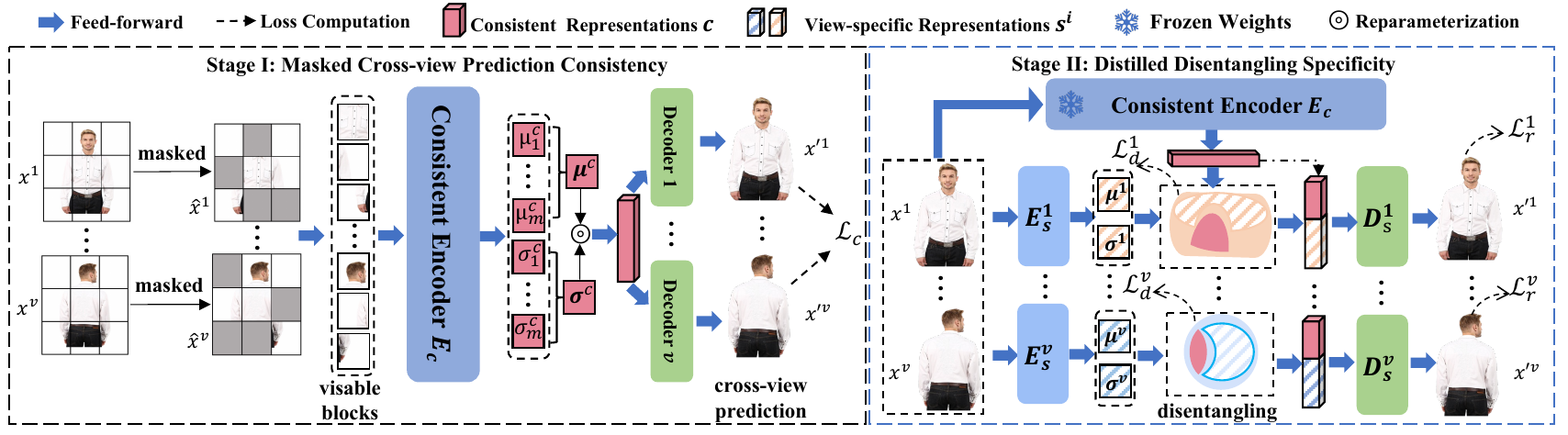}
   \caption{Illustration of the workflow of the proposed framework. The objective of stage I is to exploit the masked cross-view prediction strategy to uncover view-consistent representations. Initially, a consistent encoder is employed to learn consistent representations from all masked data. Ultimately, several decoders are utilized to predict the removed content in the corresponding views. The objective of stage II is to obtain high-quality view-specific representations by filtering out consistency-related information in specific representations. We assume the standard Gaussian distribution as the prior for all representations.}
   \label{fig:framework}
\vspace{-10pt}
\end{figure*}

\textbf{Disentangled Representation Learning.}
Disentangled representation learning (DRL) aims to unravel independent control factors within latent representations~\cite{bengio2013representation, xie24cvpr, xie2023towards, ren2022co, ren2021learning}. As DRL encompasses a vast area, detailed insights can be found in a comprehensive survey by Wang et al.~\cite{wang2022disentangled}. Here, we concentrate on DRL specifically within the realm of multi-view learning. Multi-view DRL can be broadly classified into two streams: model-based disentangled methods~\cite{ke2021conan, xu2022multi, trosten2021reconsidering, tian2020contrastive} and information-theoretic disentangled methods~\cite{federici2020learning, ke2023disentangling, bao2021disentangled, xu2021multi}. 

Model-based disentangled methods aim to compel the model to differentiate between shared and view-specific information, often employing constraints like adversarial~\cite{zhou2020end} and contrastive constraints~\cite{ke2021conan, ke2022mori}. For instance, Zhang et al.~\cite{zhang2019ae2} developed a framework with nested autoencoders: the inner autoencoders focus on view-specific information, while the outer ones consolidate this information to extract globally shared knowledge. However, these methods are often constrained by their reliance on specific model architectures.
Information-theoretic disentangled methods, on the other hand, strive for generalization. A representative work~\cite{bao2021disentangled} utilizes the information bottleneck principle, predicting each view's data using information from other views. 


Our approach employs a distilled disentangling strategy, initially extracting consistent representations across views and subsequently decoupling each view to derive view-specific representations.

\textbf{Masked Modeling.}
The masked modeling technique is a self-supervised learning approach that involves predicting masked components based on the context. In natural language processing, BERT~\cite{kenton2019bert} employs masked sequences to enhance contextual understanding, a technique now widespread in training large language models~\cite{LanCGGSS20, liu2019roberta, baevski2020wav2vec}. In the visual domain, denoising autoencoders~\cite{vincent2008extracting} use a similar principle to learn robust representations by reconstructing corrupted parts of images. Recent studies~\cite{he2022masked, wei2022masked, li2023scaling} have demonstrated the efficacy of masked modeling in learning effective representations from large-scale vision datasets. Our work extends this concept, discovering that masked modeling is beneficial for learning shared information across different views, with a higher masked ratio enhancing model generalization. To the best of our knowledge, this is the first attempt to develop masked modeling specifically for extracting consistent representations across multiple views.

\section{Method}
\label{sec:method}
Given a multi-view dataset with $n$ samples and $v$ views $\mathcal{D}=\{ x^1, x^2, \cdots , x^v | x^i \in \mathbb{R}^{n \times d_{x^i}}\}$, where $d_{x^{i}}$ is the dimensionality of $i$-th view. Our objective is to learn high-quality view-consistent representations and view-specific representations from $\mathcal{D}$. To simplify the notation throughout this paper, the subscript denotes $j^{th}$ data sample, and the superscript $^i$ indexes the view. 

\subsection{Overall Architecture}
As shown in Fig.~\ref{fig:framework}, our approach employs the two-stage pipeline to obtain high-quality consistency and specificity in the multi-view data. In the first stage, we use the random masking technique to process multi-view data and obtain masked multi-view samples. Then, we use a consistent encoder $E_c$, to extract view-consistent representations $\bold{c}$ from the visible blocks. These view-consistent representations then serve as inputs for view-specific decoders, generating reconstructed views We refer to this process as \textit{Masked Cross-view Prediction}. In the second stage, we leverage a series of view-specific encoders $\{E_s^i\}^v_{i=1}$ to extract view-specific representations $\{\bold{s}^i\}^v_{i=1}$. After that, we use a disentangling module to minimize the mutual information (MI) between the view-consistent representations $\bold{c}$ and the view-specific representations $\{\bold{s}^i\}^v_{i=1}$. Note that, for interpretability of the representations, we require that the consistency and specificity obey the standard normal distributions $\mathcal{N}(\bold{0}, \bold{I})$. Below, we depict the details of mining consistency and specificity, respectively.

\subsection{Masked Cross-view Prediction Consistency}

Multi-view consistency encapsulates the shared information across different views. Building on this foundational concept, we deduce a critical implication: an effective view-consistent representation has the capacity to predict diverse views of the same object concurrently, devoid of supplementary information. Consequently, our objective is to obtain a representation that seamlessly encapsulates distinct views. However, the presence of view-specific content inherent to each view can hinder the extraction of shared information. To address this challenge, we employ the masked technique, randomly obscuring a portion of the original signals from multiple views, as illustrated in Fig.~\ref{fig:framework}. Subsequently, we utilize the visible blocks as input for the consistent encoder, facilitating the learning of view-consistent representations to predict each view. The intuition behind this approach is that random masking diminishes view-specific information, compelling the consistent encoder to discern consistency in the visible segments. Moreover, the masked prediction can be viewed as a denoising process, thereby enhancing the robustness of view-consistent representations. In practical applications, the choice of the consistency encoder determines its approach. If based on Transformers \cite{vaswani2017attention}, it can directly handle visible blocks using patch techniques. Conversely, if employing convolutional networks \cite{he2016deep}, it can directly process the masked data.

To enhance the interpretability of consistency, we assume that the prior for view-consistent representations is standard Gaussian distribution, i.e., $p(\bold{c}) \sim \mathcal{N}(\bold{0}, \bold{I})$. Therefore, the approximate posterior $q_{\phi}(\bold{c}|\{x^i\}^v_{i=1})$ can be expressed as:
\begin{equation}
    \label{eq:consist-poster}
    q_{\phi}(\bold{c}|\{x^i\}^v_{i=1}) = \prod_{i=1}^{v}p_{\phi}(\bold{c} | x^i),
\end{equation}
where $\phi$ is the trainable parameters of consistent encoder $E_c(\cdot)$. However, Eq. (\ref{eq:consist-poster}) is intractable to optimize. To address it, we can use the reparameterization trick \cite{KingmaW13} to re-write the equality:
\begin{equation}
    \label{eq:repara-trick}
    q_{\phi}(\bold{c}|\{x^i\}^v_{i=1}) = \mathcal{N}(\bold{\mu^{c}}, (\bold{\sigma^{c}})^2) = \bold{\mu^{c}} + \bold{\sigma^{c}}\bold{\epsilon^{c}},
\end{equation}
where $\bold{\epsilon^c} \sim \mathcal{N}(0, 1)$. $\mu^{c}$ and $\sigma^{c}$ are parameterized with neural networks. For convenience, we can directly employ vanilla variational autoencoders (VAEs) \cite{KingmaW13} to construct MCP. To further ensure that the consistent encoder can learn multi-view consistency, we adopt a series of decoders to store view-specific content. Overall, we can derive the evidence lower bound (ELBO) of multi-view consistency:
\begin{equation}
\begin{aligned}
    \label{eq:consist-elbo}
    \mathcal{L}_c =  \mathbb{E}_{q(\bold{c}|\{\hat{x}^{i}\}^v_{i=1})}&[\log p(\{x^i\}^v_{i=1} | \bold{c})] \\
    & - \bold{KL}(q_\phi(\bold{c}|\{\hat{x}^i\}^v_{i=1}) \| p(\bold{c})),
\end{aligned}
\end{equation}
where $\{\hat{x}^{i}\}^v_{i=1}$ is the masked version of $\{x^i\}^v_{i=1}$. $\bold{KL}(\cdot)$ is the KL divergence. The purpose of Eq. (\ref{eq:consist-elbo}) is to learn effective view-consistent representations by minimizing the gap between masked data and original data while making the posterior of consistency $q(\bold{c}|\{x^i\}^v_{i=1})$ try to match its prior $p(\bold{c})$. During the inference stage, we discard the view-specific decoders and solely utilize the consistent encoder $E_c(\cdot)$ to obtain the consistent representations, i.e., $\bold{c} = E_c(\{x^i\}^{v}_{i=1})$. The consistent encoder can accept both masked inputs and original inputs in the inference process.

\subsection{Distilled Disentangling Specificity}

View-specific content typically encompasses specific semantic details within a view, including angles, distances, and positional information. In general, a view results from the amalgamation of both view-specific and view-consistent information through intricate higher-order interactions. However, deducing these underlying higher-order interactions from observational data poses a formidable challenge. Consequently, the objective of this section is to disentangle corresponding view-specific information from diverse views in an environment characterized by unknown interactions.

We having obtained view-consistent representations in the previous section, a natural idea is to leverage this known factor to explore the remaining information in multi-views. We denote this approach as \textit{Distilled Disentangling} (DD). Specifically, we employ the consistent encoder trained in the previous stage to generate the view-consistent representations. Then, we employ a series of view-specific encoders $\{E_s^i(\cdot)\}^v_{i=1}$ to obtain coarse view-specific representations for each view. Following this, we employ the DD module to obtain fine-grained view-specific representations $\{\bold{s}^i\}^v_{i=1}$, by minimizing the upper bound of mutual information between coarse view-specific representations and view-consistent representations. Lastly, we concatenate the consistent representations and view-specific representations, i.e., $z^i=[\bold{c}, \bold{s}^i]$, serving as the input to the view-specific decoder $D_s^i(\cdot)$. The intuition behind this strategy is that, once view-specific representations are obtained through the DD module, we can employ the decoders to simulate higher-order interactions. This implies that the higher the reconstruction quality, the higher the quality of the disentangled representations we obtain.

Based on the previous description, we also assume that the prior of the specificity variable is a standard normal distribution, i.e., $p(s^i) \sim \mathcal{N}(\bold{0}, \bold{I})$. Similar to the process of modeling consistency, we have $g_{\theta^i}(s^i|x^i)$ as the posterior for specificity, where $\theta^i$ is the trainable parameters of the view-specific encoder $E_s^i(\cdot)$. Similar to Eq. (\ref{eq:repara-trick}), we also have the equality: $g_{\theta^i}(s^i|x^i) = \mu^{i} + \sigma^{i}\epsilon^{i}$. To obtain disentangled view-specific representations, we need to minimize the mutual information upper bound between $s^i$ and $\bold{c}$:
\begin{equation}
\begin{aligned}
\label{eq:upper-bound}
I(s^i; \bold{c}) & \equiv \mathop{\mathbb{E}}\limits_{(s^i, \bold{c})\sim p_{s^i, \bold{c}}(\cdot)}[\log \frac{p(s^i |\bold{c})}{p(s^i)}] \\
&= \mathop{\mathbb{E}}\limits_{(s^i, \bold{c})\sim p_{s^i, \bold{c}}(\cdot)}[\log \frac{p(s^i | \bold{c})}{g(s^i)}] - \bold{KL}[p(s^i) \| g(s^i)] \\
& \leq \mathop{\mathbb{E}}\limits_{(s^i, \bold{c})\sim p_{s^i, \bold{c}}(\cdot)}[\log \frac{p(s^i | \bold{c})}{g(s^i)}] \\
&= \bold{KL}[p(s^i | \bold{c}) \| g(s^i)],
\end{aligned}
\end{equation}
where $g(\cdot)$ is a variational marginal approximation. This inequality has two limitations: i) it necessitates non-negativity of the KL-divergence, and ii) learning a good marginal approximation $g(s^i)$ to match $p(s^i)$ is a challenge. To address it, we introduce the CLUB estimator \cite{cheng2020club} which use a variational distribution $q_{\theta^i}(s^i | \bold{c})$ to approximate $p(s^i | \bold{c})$, more details given by \cite{cheng2020club}:
\begin{equation}
\label{eq:club}
\begin{aligned}
\mathcal{L}_d^i = I_{\bold{CLUB}}(s^i; \bold{c})	& := \mathop{\mathbb{E}}\limits_{(s^i, \bold{c})\sim p_{s^i, \bold{c}}(\cdot)}[\log q_{\theta^i}(s^i | \bold{c})] \\
& - \mathop{\mathbb{E}}\limits_{\bold{c}\sim p_{\bold{c}}(\cdot)}\mathop{\mathbb{E}}\limits_{s^i\sim p_{s^i}(\cdot)}[\log q_{\theta^i}(s^i | \bold{c})] \\
& \geq I(s^i; \bold{c}).
\end{aligned}
\end{equation}
The benefit of Eq. (\ref{eq:club}) is that we can obtain disentangled representations indirectly by minimizing $\mathcal{L}_d$, which equally minimize the mutual information upper bound between $s^i$ and $\bold{c}$. Additionally, we can conveniently estimate $q_{\theta^i}(s^i | \bold{c})$ using neural networks.

To prevent the model obtaining trivial solutions during the disentangling process, we concatenate all the disentangled representations into a new vector $z^i = [s^i, \bold{c}]$, and then input $z^i$ into the corresponding view-specific decoder $D_s^i(\cdot)$ to obtain the reconstruction data, i.e., $x^{'i} = D_s^i(z^i)$. The loss function for this process is as follows:
\begin{equation}
\label{eq:recon-loss}
\begin{split}
	\mathcal{L}_{r}^{i} = \ & \mathop{\mathbb{E}}\limits_{\mathbf{z}^i \sim q_{\theta^i}(\mathbf{z}^i | x^i)}[\log p (x^i | \mathbf{z}^i) ]  - \mathbf{KL} [q_{\theta^i}(\mathbf{z}^i | x^i) \| p(\mathbf{z}^i) ].
\end{split}	
\end{equation}
Eventually, our loss function of the second stage contains two parts:
\begin{equation}
\label{eq:stage2-loss}
\mathcal{L}_s = \frac{1}{v} \sum_{i=1}^{v} \mathcal{L}_d^i + \mathcal{L}_{r}^{i}
\end{equation}
where the first term is optimized to learn disentangled view-specific representations. The second term is optimized to maintain the reconstruction quality. The pseudo-code of MRDD is given in the supplement.



\section{Experiments}

\subsection{Dataset}
\label{sec:dataset}
We evaluate the proposed method and other competitive methods using five multi-view datasets. There are: \textbf{(a) E-MNIST} \cite{liu2016coupled}, which is a well-known benchmark dataset consisting of 70,000 grayscale digit images (0-9) with $32 \times 32$ pixels. The views contain the original digits and the edge-detected version, respectively; \textbf{(b) E-FMNIST \cite{xiao2017fashion}}, which is a fashion dataset consisting of $32 \times 32$ grayscale images of clothing items. We synthesize the second view by running the same edge detector used to create E-MNIST; \textbf{(c) COIL-20 \cite{nene1996columbia}} and \textbf{(d) COIL-100 \cite{nene1996columbia}} which depicts from different angles containing grayscale images of 20 items and RGB images of 100 items, respectively. We create a three-view dataset by randomly grouping the images for an item into groups of three; \textbf{(f) Office-31} \cite{saenko2010adapting}, which consists of objects commonly encountered in office settings, such as keyboards, file cabinets, and laptops. We leverage the ColorJitter method to construct a three-view dataset. We report the dataset description in the supplement.



\subsection{Baseline Models}\label{sec:baseline-models}
We elvaluate two versions of the proposed method: MRDD-$\bold{c}$ represents the use of only view-consistent representations $\bold{c}$, while MRDD-$\bold{cs}$ denotes the use of $\boldsymbol{c}$ and the first view-specific representations $s^1$. We set up three categories of baseline models for comparison with MRDD, including \textbf{(i) single-view disentangling methods:} Joint-VAE \cite{dupont2018learning} and $\beta$-VAE~\cite{higgins2016beta}; \textbf{(ii) model-based multi-view disentangling methods:} MFLVC~\cite{xu2022multi}, EAMC~\cite{zhou2020end}, CONAN~\cite{ke2021conan}, CMC~\cite{tian2020contrastive}, and GCFAgg \cite{yan2023gcfagg}; and \textbf{(iii) information-theory-based multi-view disentangling methods:} Multi-VAE~\cite{xu2021multi}, MIB~\cite{federici2020learning}, DVIB~\cite{bao2021disentangled}, and UNITER~\cite{xu2023untie}. Our method falls into the third category. We evaluate the clustering and classification performance of all comparative models using k-means and support vector classification (SVC). Each model undergoes 10 runs, and we report their average values and variances. Notably, for single-view methods, we select the results from the best view as the evaluation outcome. In the case of multi-view methods limited to two views, we choose the optimal two views as inputs for these models, such as MIB and DVIB.

\begin{table*}[t]
\setlength{\abovecaptionskip}{0cm} 
\setlength{\belowcaptionskip}{-0.2cm} 
\setlength\tabcolsep{2pt}
\begin{center}
\scalebox{0.85}{
\begin{tabular}{lcccccccccc}\toprule

 & \multicolumn{2}{c}{E-MNIST} & \multicolumn{2}{c}{E-FMNIST} & \multicolumn{2}{c}{COIL-20} & \multicolumn{2}{c}{COIL-100} & \multicolumn{2}{c}{Office-31} \\
\cmidrule(lr){2-3} \cmidrule(lr){4-5} \cmidrule(lr){6-7} \cmidrule(lr){8-9} \cmidrule(lr){10-11} 
Method & ACC$_{clu}$ & NMI & ACC$_{clu}$ & NMI & ACC$_{clu}$ & NMI & ACC$_{clu}$ & NMI & ACC$_{clu}$ & NMI \\
\midrule
\rowcolor{Gray}Joint-VAE \cite{dupont2018learning} & 42.81$\pm$0.03 & 35.45$\pm$0.05 &  37.22$\pm$0.58 &  26.94$\pm$0.31 & 61.98$\pm$2.60 & 74.14$\pm$0.82 & 55.85$\pm$1.49 & 77.53$\pm$0.61 & 25.19$\pm$0.60 & 29.30$\pm$0.34   \\
$\beta$-VAE \cite{higgins2016beta} & 39.69$\pm$0.72  & 24.97$\pm$0.18 & 39.76$\pm$0.02 & 38.37$\pm$0.05 & 35.80$\pm$0.71 & 44.67$\pm$0.86 & 24.02$\pm$0.23 & 40.96$\pm$0.26 &  11.89$\pm$0.44 &  13.53$\pm$0.09   \\
\midrule
\rowcolor{Gray}MFLVC$\dag$~\cite{xu2022multi} & 65.98$\pm$0.06 & 59.08$\pm$0.04 & 48.48$\pm$0.15 & 46.62$\pm$0.12 & 36.98$\pm$2.07 & 67.16$\pm$1.99  & 35.03$\pm$1.14 & 73.19$\pm$1.16 & 32.09$\pm$0.21 & 29.39$\pm$0.07 \\
CONAN$\dag$ \cite{ke2021conan} & 50.22$\pm$0.02 & 44.42$\pm$0.01 & 48.70$\pm$0.11 & 41.41$\pm$0.01 & 55.94$\pm$0.88 & 63.98$\pm$0.46  & 54.04$\pm$1.37 & 74.73$\pm$0.47 & 13.51$\pm$0.24 & 17.11$\pm$0.22 \\
\rowcolor{Gray}CMC$\dag$ \cite{tian2020contrastive} & 64.16$\pm$1.04 & 59.84$\pm$0.77 & 50.13$\pm$0.42 & 46.24$\pm$0.41 & 58.03$\pm$1.18 & 71.23$\pm$1.22 & 57.19$\pm$0.94 & 78.16$\pm$1.07 & \underline{34.28$\pm$0.32} & 29.75$\pm$0.19 \\
EAMC$\dag$ \cite{zhou2020end} & 49.17$\pm$0.32 & 46.28$\pm$0.34 &  45.44$\pm$1.08 & 42.76$\pm$1.03 & 58.19$\pm$1.93 & 75.13$\pm$1.21 & \underline{60.31$\pm$0.84} & 73.13$\pm$0.91 & 33.16$\pm$0.19 & 30.08$\pm$0.13 \\
\rowcolor{Gray}GCFAgg$\dag$ \cite{yan2023gcfagg} & \underline{67.10$\pm$0.88} & \underline{61.34$\pm$0.62} & 43.09$\pm$0.07 & 40.25$\pm$0.21 & 55.79$\pm$1.66 & 75.08$\pm$1.38  & 45.71$\pm$1.35 & 70.22$\pm$1.82 & 31.17$\pm$0.21 & 28.54$\pm$0.31 \\
\midrule
Multi-VAE \cite{xu2021multi} & 60.74$\pm$0.23 & 59.03 $\pm$0.18 & 53.16$\pm$0.14 & \underline{54.47$\pm$0.06} & 65.77$\pm$1.04 & 78.22$\pm$1.03 &  48.87$\pm$0.03 & 45.29$\pm$0.15 & 31.27$\pm$0.27 & 27.84$\pm$0.39\\
\rowcolor{Gray}MIB \cite{federici2020learning} & 53.16$\pm$0.43 & 48.06$\pm$1.03 & \underline{53.71$\pm$0.72} & 52.66$\pm$0.59 & 53.60$\pm$0.89 & \underline{80.61$\pm$0.76} & 53.27$\pm$0.13 & \underline{81.53$\pm$1.05} & 33.08$\pm$0.26 & 29.94$\pm$0.35 \\
DVIB \cite{bao2021disentangled} & 47.80$\pm$0.01 & 33.29$\pm$0.02 & 30.23$\pm$0.85 & 43.36$\pm$0.30 & 57.47$\pm$1.55 & 68.27$\pm$0.95 & 36.75$\pm$0.48 & 25.54$\pm$0.36 & 29.24$\pm$0.38 & 28.69$\pm$0.21 \\
\rowcolor{Gray}UNITER \cite{xu2023untie} & 65.35$\pm$0.11 & 59.71 $\pm$0.25 & 53.07$\pm$0.83 & 51.19$\pm$0.10 & \underline{67.54$\pm$0.99} & 79.60$\pm$1.41 &  50.12$\pm$0.44 & 54.77$\pm$0.33 & 34.40$\pm$0.02 & \underline{30.22$\pm$0.07} \\
\midrule
MRDD-$\bold{c}$ (Ours)  & 70.34$\pm$0.05 & 63.49$\pm$0.02 & 50.11$\pm$0.09 & 53.55$\pm$0.08 & 63.96$\pm$0.03 & 73.88$\pm$0.03 & \textbf{65.29$\pm$0.16} & \textbf{84.95$\pm$0.07} & 30.06$\pm$0.59 & 28.14$\pm$0.15 \\
\rowcolor{Gray}MRDD-$\bold{cs}$ (Ours)  & \textbf{75.93$\pm$0.12} & \textbf{69.00$\pm$0.41} & \textbf{58.25$\pm$0.27} & \textbf{59.93$\pm$0.39} & \textbf{69.18$\pm$0.44} & \textbf{81.52$\pm$0.14} & 62.00$\pm$0.27 & 83.65$\pm$1.04 & \textbf{37.14$\pm$0.98} & \textbf{39.26$\pm$0.53} \\
\midrule
$\Delta$ SOTA  & \textcolor{commentcolor}{+8.83} & \textcolor{commentcolor}{$+$7.66} & \textcolor{commentcolor}{$+$4.54} & \textcolor{commentcolor}{$+$4.86} & \textcolor{commentcolor}{$+$1.64} & \textcolor{commentcolor}{$+$0.91} & \textcolor{commentcolor}{$+$4.98} & \textcolor{commentcolor}{$+$3.34} & \textcolor{commentcolor}{$+$2.86} & \textcolor{commentcolor}{$+$9.04} \\
\bottomrule
\end{tabular}}
\end{center}
\caption{\textbf{Clustering results (\%) on five datasets.} \textbf{Bold} denotes the best results and \underline{underline} denotes the second-best. $\dag$ denotes we set the dimensionality of latent representations as 10. All results are reproduced using the officially released code.}
\label{tab:clustering-result}
\end{table*}

\subsection{Implementation Details}
We implement the proposed method and other non-linear comparison methods on the PyTorch 2.0.1 \cite{paszke2019pytorch} platform, running on Ubuntu 18.04 LTS utilizing one GPU (NVIDIA GeForce RTX 2080 Ti with 12 GB of memory). For simplicity, we use convolutional networks to build the encoders and decoders\footnote{Network structure details are given in the supplement.}. For all experiments, we default the masked ratio of MCP to $70\%$, and both the dimensions of consistency and specificity are set to 10. It is worth noting that the MCP's masked ratio is set at $70\%$ to strike a balance between minimizing information loss and optimizing encoder computation cost. We train two stages of our model for 200 epochs, respectively. We use the Adam optimizer with a batch size of 512 for the model and then employ the cosine annealing learning rate scheduler with an initial learning rate of $5\times10^{-4}$. To ensure a fair comparison, we use the recommended optimal settings from the original paper for other baseline models. In the classification task, we split the dataset into a training set and a test set in a ratio of 80:20.

\subsection{Evaluation Metrics}
In order to evaluate clustering performance, two standard evaluation metrics are used: clustering ACCuracy (ACC$_{clu}$) and Normalized Mutual Information (NMI). Readers seeking further details on these metrics are referred to \cite{kumar2011co}. It is important to note that the validation process of clustering methods is limited to cases where ground truth labels are available. For classification, ACC$_{cls}$ and F-Score are used as evaluation metrics. In all cases, a high value indicates better performance\footnote{The details of the evaluation metrics are given in the supplementary materials.}.

\begin{table}[t]
\setlength{\abovecaptionskip}{0cm} 
\setlength{\belowcaptionskip}{-0.2cm} 
\setlength\tabcolsep{2pt}
\begin{center}
\scalebox{0.62}{
\begin{tabular}{lcccccc}
\toprule
 & \multicolumn{2}{c}{E-MNIST} & \multicolumn{2}{c}{COIL-100} & \multicolumn{2}{c}{Office-31} \\
\cmidrule(lr){2-3} \cmidrule(lr){4-5} \cmidrule(lr){6-7} 
Method & ACC$_{cls}$ & F-Score & ACC$_{cls}$ & F-Score & ACC$_{cls}$ & F-Score \\
\midrule
\textit{Random} & 10.00$\pm$0.08 & 10.00$\pm$0.08 & 1.10$\pm$0.15 & 1.08$\pm$0.13 & 3.17$\pm$0.15 & 3.13$\pm$0.14 \\
\rowcolor{Gray}Joint-VAE\cite{dupont2018learning} & 81.16$\pm$0.18 & 80.97$\pm$0.20 & \underline{86.73$\pm$0.86} & \underline{85.94$\pm$0.80} & 42.32$\pm$1.10 & 40.51$\pm$1.15 \\
$\beta$-VAE \cite{higgins2016beta} & 43.82$\pm$0.38 & 41.84$\pm$0.39 & 30.87$\pm$1.15 & 25.64$\pm$0.83 &  28.09$\pm$0.75 &   25.25$\pm$0.94 \\
\midrule
\rowcolor{Gray}CONAN$\dag$ \cite{ke2021conan} & 61.42$\pm$0.36 &  59.07$\pm$0.36 &  62.45$\pm$1.30 &  57.33$\pm$0.61 & 47.07$\pm$1.47 &  44.79$\pm$1.41\\
CMC$\dag$ \cite{tian2020contrastive} & 93.86$\pm$0.04 & 93.05$\pm$0.01 & 80.29$\pm$0.51 & 79.48$\pm$1.10 & 51.33$\pm$0.46 & 50.87$\pm$0.62 \\
\rowcolor{Gray}Multi-VAE \cite{xu2021multi} & 92.73$\pm$0.16 & 92.76$\pm$0.16 & 74.11$\pm$2.18 & 70.78$\pm$0.95 & 62.53$\pm$1.17 & 62.20$\pm$1.39 \\
MIB \cite{federici2020learning} &  90.81$\pm$0.30 & 90.03$\pm$0.49 &  79.01$\pm$0.04 & 78.63$\pm$0.09 &  61.48$\pm$0.43 & 60.79$\pm$0.39    \\
\rowcolor{Gray}DVIB \cite{bao2021disentangled} & 84.14$\pm$0.19 & 83.98$\pm$0.24  & 67.22$\pm$0.73 & 66.34$\pm$1.16 & 58.12$\pm$0.87 & 56.52$\pm$0.77 \\
UNITER \cite{xu2023untie} & \underline{94.21$\pm$0.16} & \underline{94.11$\pm$0.38} & 84.26$\pm$1.05 & 84.23$\pm$1.21 & \underline{69.33$\pm$1.42} & \underline{67.51$\pm$0.99} \\
\midrule
\rowcolor{Gray}MRDD-$\bold{c}$ (Ours) & 93.69$\pm$0.17 & 93.67$\pm$0.18 & 86.67$\pm$1.65 & 84.86$\pm$1.96 & 53.04$\pm$0.83 & 48.92$\pm$1.38       \\
MRDD-$\bold{cs}$ (Ours) & \textbf{98.37$\pm$0.09} & \textbf{98.36$\pm$0.09} & \textbf{91.17$\pm$1.21} & \textbf{90.06$\pm$1.15} & \textbf{73.51$\pm$0.28} & \textbf{72.71$\pm$0.44}        \\
\midrule
$\Delta$ SOTA  & \textcolor{commentcolor}{$+$4.16} & \textcolor{commentcolor}{$+$4.25} & \textcolor{commentcolor}{$+$4.44} & \textcolor{commentcolor}{$+$4.12} & \textcolor{commentcolor}{$+$4.18} & \textcolor{commentcolor}{$+$5.2}\\

\bottomrule
\end{tabular}}
\end{center}
\caption{\textbf{Classification results (\%) on three datasets.} \textbf{Bold} denotes the best results and \underline{underline} denotes the second-best. $\dag$ denotes we set the dimensionality of latent representations as 10. All results are reproduced using the officially released code.}
\label{tab:classification-result}       
\end{table}

\subsection{Comparison Results and Analysis}

We evaluate the performance of all baseline models on five datasets for both clustering and classification tasks\footnote{More results are given in the supplement.}, as shown in Tables \ref{tab:clustering-result} and \ref{tab:classification-result}. These results indicate that, under the same experimental settings, the representations extracted by our method significantly improve the performance of downstream tasks. At the same time, we find that MRDD-$\bold{c}$ exhibits superior performance in both ACC$_{clu}$ and NMI metrics compared to MRDD-$\bold{cs}$. This suggests that naively concatenating features may lead to a degradation in the performance of the representation. In addition, the results on COIL-20 and COIL-100 indicate that Joint-VAE outperforms the majority of multi-view methods, such as MFLVC, CMC, and GCFAgg, in terms of evaluation metrics. This outcome highlights that the quality of views can impact the quality of representations, consequently leading to a degradation in the performance of fused representations.

\subsection{Ablation Study}

\noindent \textbf{Components study.} We conducted ablation experiments to evaluate the effects of different components in the proposed method MRDD-$\bold{cs}$, as shown in Table~\ref{tab:components-ablation}. In the case of using only Stage I and Stage II respectively indicate that our framework is superior to any of the two independent components. Moreover, the results of using only Stage II support our hypothesis that it is difficult to extract high-quality multi-view representations in an unsupervised environment without additional information. On the other hand, the ablation results show that removing MCP will degrade the performance of our method. Notably, the ablation experiments of $\mathcal{L}_d$ and $\mathcal{L}_r$ demonstrate that removing $\mathcal{L}_r$ has a more significant negative impact on our method.

\begin{table}[t]
\setlength{\abovecaptionskip}{0.4cm} 
\setlength{\belowcaptionskip}{-0.2cm}
\setlength\tabcolsep{2pt}

\centering
\scalebox{0.75}{
\begin{tabular}{llllll}
\toprule
Method & E-MNIST & E-FMNIST & COIL-20 & COIL-100 & Office-31 \\
\midrule
MRDD-$\bold{cs}$ (baseline) & \textbf{98.37} & \textbf{88.78} & \textbf{95.97} & \textbf{91.17} & \textbf{73.51} \\
\quad only stage I & 93.69 & 82.51 & 88.18 & 86.67 & 53.04  \\ 
\quad only stage II & 62.13 & 58.64 & 69.65 & 61.28 & 40.68 \\
\quad w/o MCP & 93.12  & 83.11  & 89.48 & 86.93 & 61.45  \\
\quad w/o $\mathcal{L}_d$ & 94.57 & 83.07  & 90.72  & 85.37 & 58.92  \\
\quad w/o $\mathcal{L}_r$ & 90.63  & 76.88 & 86.90  & 88.96  & 50.04 \\ 
\bottomrule
\end{tabular}
}
\caption{\textbf{Components ablation study.} Classification accuracy scores (\%) on five dataset. Best results are in \textbf{bold}.}
\label{tab:components-ablation}       
\vspace{-10pt}
\end{table}

\begin{figure}[t]
\setlength{\abovecaptionskip}{0cm} 
\setlength{\belowcaptionskip}{-0.2cm}
    \centering
    \includegraphics[width=1\linewidth]{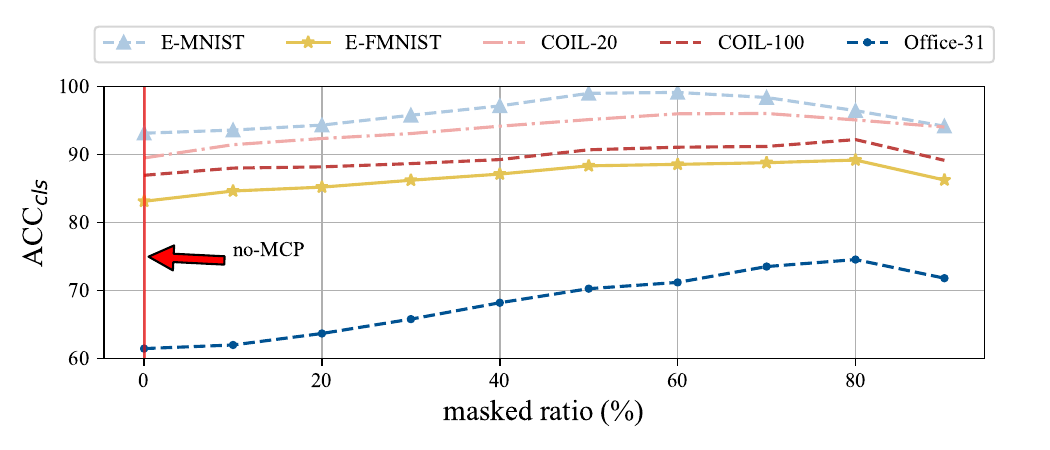}
   \caption{\textbf{Masked ratio.} Classification accuracy scores (\%) for masked ratios range from $0\%$ to $90\%$ on five datasets.}
   \label{fig:masked-ratio}
\end{figure}

\noindent \textbf{Masked ratio and strategies.} In this study, we conduct a series of ablation studies focusing on the MCP component, applying various masked ratios and masking strategies. The results of Fig.~\ref{fig:masked-ratio} show a notable trend: an increase in the masked ratio correlates positively with the model's classification performance. It is observed that, for the majority of the datasets examined, an optimal performance is attained at a masked ratio of $70\%$. This observation aligns well with the findings reported in prior studies. Interestingly, specific datasets, notably COIL-20 and Office-31, exhibited a peak in classification accuracy at a high masked ratio of $80\%$. Furthermore, it was observed that even at a masked ratio of $90\%$, the decrement in performance was not pronounced. This resilience in performance at high masked ratios can be postulated to stem from the multi-view context employed, wherein the model retains the capability to assimilate valuable representations from the visible blocks of alternate views.

Furthermore, our research included an exploration of three distinct masking strategies—random, block-wise, and grid-wise—to evaluate their respective impacts on the model's performance. By default, our methodology utilized the random strategy. The comparative analysis of these strategies, as depicted in Table~\ref{tab:masked-strategy}, indicates that the differences in performance metrics among the three strategies are relatively marginal. Notably, the random strategy outperformed the others across all datasets. The performance of the block-wise strategy was observed to be inferior compared to the other two. We conjecture that this underperformance may be attributed to the block-wise strategy's inherent design, which involves the removal of large blocks, potentially leading to the concurrent loss of shared information.

\begin{table}[t]
\setlength{\abovecaptionskip}{0cm} 
\setlength{\belowcaptionskip}{-0.2cm}
\begin{center}
\scalebox{0.70}{
\begin{tabular}{lccccc}
\toprule
Strategy & E-MNIST & E-FMNIST & COIL-20 & COIL-100 & Office-31 \\
\midrule
Random & \textbf{98.37} & \textbf{88.78} & \textbf{95.97} & \textbf{91.17} & \textbf{73.51} \\
Block-wise \cite{Bao0PW22} & 96.90 & 87.24 & 94.55 & 89.79 & 70.14 \\
Grid-wise \cite{he2022masked} & 97.15 & 87.62 & 93.19 & 90.03 & 71.43 \\
\bottomrule
\end{tabular}
}
\end{center}
\caption{\textbf{Masked strategy.} Classification results (\%) of three masked strategies (e.g., random, block-wise, and grid-wise) with the masked ratio of $70\%$ on five datasets. Best results are in \textbf{bold}.}
\label{tab:masked-strategy}       
\end{table}

    

\begin{figure}[t]
\setlength{\abovecaptionskip}{0.3cm} 
\setlength{\belowcaptionskip}{-0.2cm}
    \centering
    \includegraphics[width=1\linewidth]{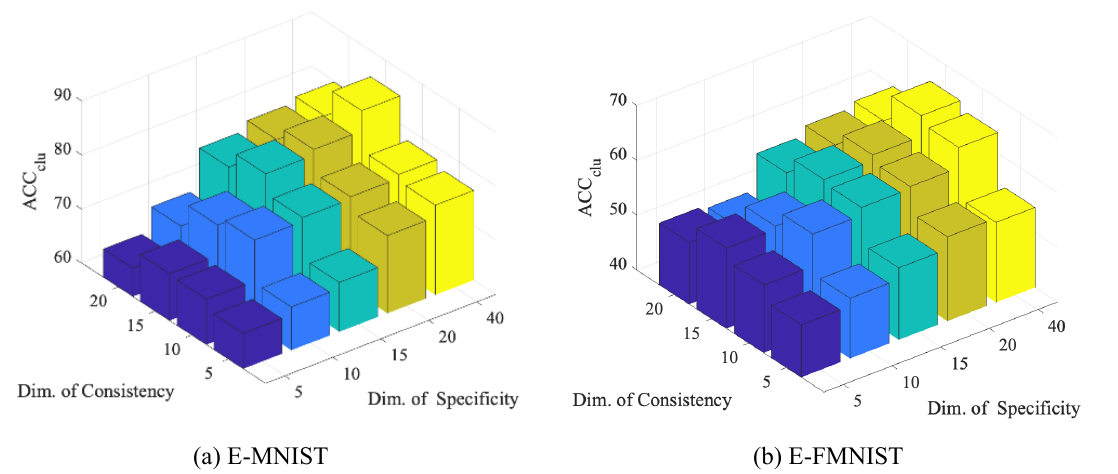}
   \caption{The clustering results (\%) of the different dimensions of consistency and specificity on the E-MNIST and E-FMNIST datasets. The x-axis represents the consistency dimension, the y-axis represents the specificity dimension, and the z-axis represents the clustering accuracy.}
   \label{fig:dimension}
\end{figure}

\noindent \textbf{The dimensionality of consistency and specificity.} We study the effects of view-consistent and view-specific representations extracted by our method in different dimensions\footnote{More results are given in the supplement.}. We set the range of view-consistent representation dimensions to $\{5, 10, 15, 20\}$ and the view-specific representation dimension to $\{5, 10, 15, 20, 40\}$. Fig.~\ref{fig:dimension} shows that when the consistent representation dimension is fixed, the clustering performance is positively correlated with the view-specific representation dimension. Our method achieved the best clustering performance on the E-MNIST and E-FMNIST datasets when the consistent representation dimension is 15 and the specific representation dimension is 40. On the other hand, Fig.~\ref{fig:dimension} indicates that a large dimension of consistent representations can lead to performance degradation. For example, when the consistent representation dimension is set to 20, the clustering results were inferior to the consistent representation dimension of 15.

\noindent \textbf{The contributions of consistency and specificity.} We evaluate the clustering results of view-consistent and view-specific representations on five datasets. From Table~\ref{tab:contribute-repre}, we note that the manner in which features are combined exerts a pronounced effect on the performance of downstream tasks. This is attributable to the fact that we have generated additional views by applying data augmentation to a single view, resulting in the first view containing a greater quantity of information. Consequently, the use of just the $\bold{c}+\bold{s}^{1}$ feature combination can achieve superior performance. On the other hand, we also observe that a simple concatenation of all features introduces redundancy, which in turn diminishes the performance of downstream models. These experimental findings suggest that a singular approach to feature fusion is unlikely to have a universally positive impact on all tasks.


\begin{table}[t]
\setlength{\abovecaptionskip}{0cm} 
\setlength{\belowcaptionskip}{-0.2cm}
\begin{center}
\scalebox{0.75}{
\begin{tabular}{lccccc}
\toprule
Type & E-MNIST & E-FMNIST & COIL-20 & COIL-100 & Office-31 \\
\midrule
$\bold{c}$& 70.34 & 50.11 & 63.96 & \textbf{65.29} & 30.06 \\
$\bold{s}^{1}$  & 55.93 & \textbf{59.73} & 63.14 & 60.29 & 34.42 \\
$\bold{s}^{2}$  & 43.32 & 59.40 & 63.44 & 59.92 & 32.51 \\
$\bold{s}^{3}$  & - & - & 63.19 & 61.46 & 33.80 \\
$\bold{c}+\bold{s}^{1}$  & \textbf{75.93} & 58.25 & \textbf{69.18} & 62.00 & 37.14 \\
concat. & 73.16 & 58.10 & 63.26 & 60.90 & \textbf{40.84} \\
\bottomrule
\end{tabular}
}
\end{center}
\caption{The clustering results (\%) of different types of representations on five datasets, where $\bold{c}$ represents view-consistent representations, $\bold{s}^{i}$ represents the $i$-th view-specific representations, and \textit{concat}. represents concatenating all representations. Best results are in \textbf{bold}.}
\label{tab:contribute-repre}       
\vspace{-10pt}
\end{table}

\subsection{Visualization}

We visualize the representations of MRDD-$\bold{c}$ and MRDD-$\bold{cs}$ on the E-MNIST and E-FMNIST datasets\footnote{More visualization results are given in the supplement.}. Fig.~\ref{fig:tsne} indicates that view-consistent representations can distinguish different samples at a coarse level. However, after incorporating view-specific representations, the discriminative ability of the representations is enhanced, especially evident in the E-MNIST dataset.

\begin{figure}[t]
\setlength{\abovecaptionskip}{0.2cm} 
\setlength{\belowcaptionskip}{-0.2cm}
    \centering
    \includegraphics[width=1\linewidth]{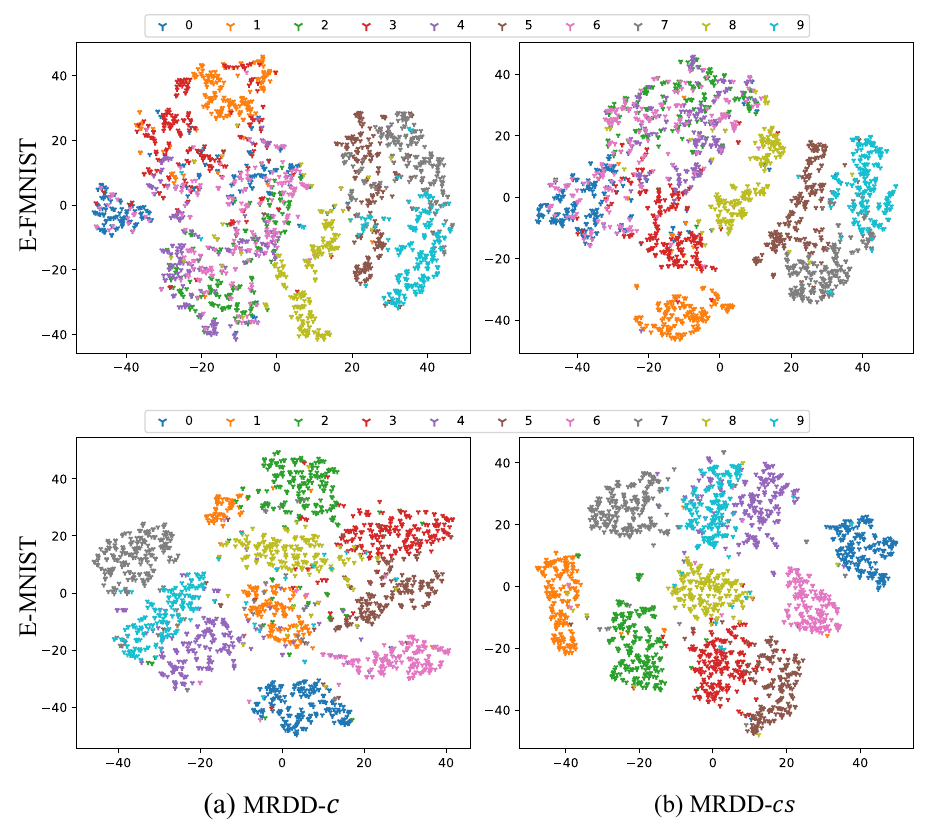}
   \caption{Visualization of the representations of MRDD-$\bold{c}$ and MRDD-$\bold{cs}$ using t-SNE~\cite{van2008visualizing} on the E-MNIST and E-FMNIST datasets.}
   \label{fig:tsne}
\vspace{-10pt}
\end{figure}


\section{Discussion and Conclusion}
\label{sec:conclusion}

This work attempts to address the problem of obtaining high-quality multi-view consistent and specific representations from the perspective of representation disentangling. Extensive experimental results reveal that: i) A high masked ratio can enhance the performance of view-consistent representations while reducing computational costs. ii) Diminishing the size of the view-consistent representations in relation to view-specific representations substantially enhances the efficacy of their integrated representations. iii) The role of view-specific representations in predictive tasks is significant, and the mere aggregation of all representations might lead to suboptimal performance. 

Although the proposed method is adept at disentangling consistency and specificity, the efficacy of this decoupling is contingent upon the quality of the consistent representations. On the other hand, while the MCP enhances performance and efficiency, it also increases the instability of training. In complex scenarios, the consistent representations obtained by MCP make it challenging to reconstruct the original data. Our future work will consider how to reduce the dependency between the two stages and improve the stability of the MCP module.

\textbf{Acknowledgment.}
This work was supported by the Beijing Natural Science Foundation (No. 4234086); the Natural Science Foundation of China (No. 62192782); Guangdong Natural Science Funds for Distinguished Young Scholar (No. 2023B1515020097); Singapore MOE Tier 1 Funds (MSS23C002); and the NRF Singapore under the AI Singapore Programme (No. AISG3-GV-2023-011).

\section*{Appendix}
\appendix
\section{Observation Experiments}
We utilized the Mutual Information Neural Estimator (MINE)\footnote{Code is accessible at: \url{https://github.com/gtegner/mine-pytorch/}}~\cite{belghazi2018mutual} as a mutual information estimator to independently assess the mutual information between view-consistent representations and view-specific representations proposed by CONAN\footnote{Code is accessible at: \url{https://github.com/Guanzhou-Ke/conan}}~\cite{ke2021conan}, DVIB\footnote{Code is accessible at: \url{https://github.com/feng-bao-ucsf/DVIB}}~\cite{bao2021disentangled}, Multi-VAE\footnote{Code is accessible at: \url{https://github.com/SubmissionsIn/Multi-VAE}}~\cite{xu2021multi}, and our approach. To ensure a fair comparison, we standardized the representation dimensions of all comparative methods to 10. For constructing the MINE estimator, we employed fully connected layers with Rectified Linear Unit (ReLU) activation, specifying the network architecture as 20-100-100-100-1. We use Adam with the learning rate of $1\times10^{-4}$ and the batch size of 128 to train the model for 500 epochs. To mitigate randomness, we executed the MINE procedure 10 times and recorded the average results.

\section{Related Work}
\label{sec:related-work}

\textbf{Multi-view Representation Learning.}
The goal of MvRL is to extract both shared and view-specific information from multiple data sources, integrating them into a cohesive representation that is advantageous for predictive tasks~\cite{LiYZ19, sun2013multi, chao2016consensus}. Existing approaches in this field generally fall into three categories: statistic-based, deep learning-based, and hybrid methods.

Statistic-based methods, employing techniques like canonical correlation analysis \cite{rasiwasia2010new, dalal2005histograms}, non-negative matrix factorization \cite{liu2013multi, wang2018multiview}, and subspace methods~\cite{brbic2018multi, wang2019multi}, excel in deriving interpretable models. However, they struggle with datasets that are high-dimensional or large-scale. In contrast, deep learning-based methods have gained prominence, especially in unsupervised settings, where generative models such as autoencoders~\cite{andrew2013deep, wang2015deep, zhang2019ae2} and generative adversarial networks~\cite{zhou2020end} are used to learn latent representations. Although effective, these methods face the challenge of redundancy when concatenating representations from all views, leading to suboptimal results for downstream tasks. Researchers have attempted to address this by exploring fusion methods for multi-view representations~\cite{ke2021conan, trosten2021reconsidering, xu2023progressive}. Nevertheless, deep learning-based methods often lack interpretability, being perceived as ``black-box'' approaches.
Hybrid methods, such as those found in \cite{zhao2017multi, huang2021deep, sun2019self}, combine statistical and deep learning approaches. They use deep learning for feature extraction and statistical learning for modeling interpretable representations. These methods effectively balance the strengths of both approaches but require substantial computational resources for post-processing.

Our approach is categorized under deep learning-based methods. We distinguish our work by utilizing deep learning's capacity to handle large datasets effectively. Moreover, we address the interpretability challenges in representations by incorporating disentanglement techniques.

\section{Pseudo-code of MRDD}

See Algorithm 1.

\begin{algorithm}[t]
	\caption{The pseudo-code of the proposed method.} 
	\label{alg1} 
	\renewcommand{\algorithmicrequire}{\textbf{Input:}}
	\renewcommand{\algorithmicensure}{\textbf{Output:}}
	\begin{algorithmic}[1]
	    \REQUIRE $\mathcal{X}=\{ x^{(1)}, x^{(2)}, \cdots , x^{(v)} | x^{(i)} \in \mathbb{R}^{n \times d_v}\}$, the consistent encoder $E_{c}$, view-specific encoders and decoders $\{E_s^{(i)}\}_{i=1}^{v}$, $\{D_s^{(i)}\}_{i=1}^{v}$
	    \ENSURE the view-consistent representation $c$, and view-specific representations $\{s^{(i)}\}_{i=1}^{v}$
	    \STATE masked inputs $\{x^{(i)}\}_{i=1}^{v} \to $ $\{\hat{x}^{(i)}\}_{i=1}^{v}$.
		\STATE $c \gets $ concatenating all of $ E_{c} (\hat{x})$ 's outputs.
		\STATE computing the consistent loss $\mathcal{L}_{c}$ using Eq.(3) .
		\STATE fixed the the consistent encoder $E_{c}$.
	    \REPEAT
	    \STATE $\{s^{(i)}\}^V_{i=1} \gets$ $E_s^{(i)}(\{x^{(i)}\}^v_{i=1})$, and $c \gets$ $E_{c}(\{x^{(i)}\}^v_{i=1})$
	    \STATE computing the disentangling loss $\mathcal{L}_{d}^{i}$ using Eq. (5).
             \STATE computing the reconstruction loss $\mathcal{L}_{r}^{i}$ using Eq. (6).
	    \UNTIL $\mathcal{L}_{s}$ convergence.
	\end{algorithmic}
\end{algorithm}

\section{Network Structures}

\begin{figure}[t]
\setlength{\abovecaptionskip}{0.3cm} 
\setlength{\belowcaptionskip}{-0.2cm}
    \centering
    \includegraphics[width=1\linewidth]{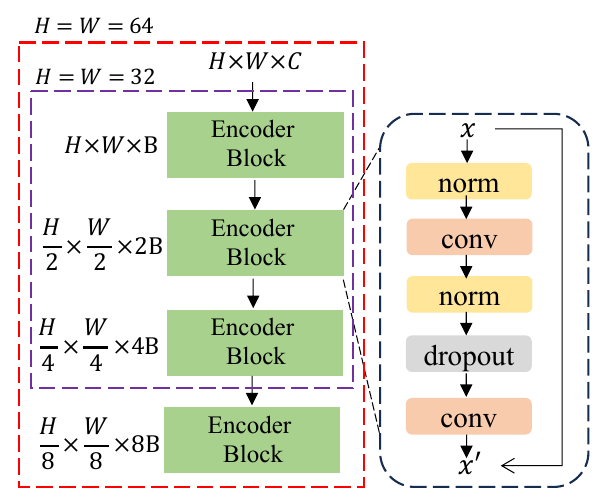}
   \caption{Illustration of encoder, where $H$, $W$, and $C$ denote the height, width, and channels of an image, respectively. $B$ denotes the number of output channels.}
   \label{fig:autoencoder}
\end{figure}

We employed convolutional neural networks to construct both the encoder and decoder components in our approach, ensuring a symmetric structure for both. As depicted in Fig.~\ref{fig:autoencoder}, an encoder block comprises two convolutional layers, two batch normalization layers, and a dropout module. These encoder blocks are then stacked to form the complete encoder. In the decoder architecture, the \texttt{Conv} module is substituted with the \texttt{ConvTranspose2d} module in PyTorch.

The output channel base, denoted as $B$, is set at 16 by default. To maintain consistent latent representations, we devised two distinct architectures tailored to different data dimensions. For data with a dimension of 32, only the first three layers from Figure 1 are utilized in both the encoder and decoder structures. Conversely, for data with a dimension of 64, we incorporate four blocks to constitute the encoder and decoder structures, ensuring the output dimension is normalized to $8\times8$. This approach is crucial for maintaining coherence across varying data dimensions.

\section{Evaluation Metrics}

To evaluate the performance of clustering, we apply three well-known metrics to the comparative experiments, including clustering accuracy (ACC$_{clu}$) and normalized mutual information (NMI). Given sample $x_j \in \mathbf{x}^i$ for any $j \in \{1,2,\cdots,n\}$, the predicated clustering label and the real label are indicated as $y_j$ and $c_j$, respectively. The ACC$_{clu}$ is defined as:
\begin{equation}
\label{eq:clustering-acc}
	ACC_{clu} = \frac{\sum_{i=1}^{N} \delta(y_j, map(c_j))}{N}
\end{equation}
where $y_j \in \mathbf{Y}$ represents ground-truth labels and $c_j \in \mathbf{C}$ denotes predicted clustering labels which generated by kmeans; $\delta(a, b)$ is the indicator function, i.e., $\delta(a, b) = 1$ if $a = b$, and $\delta(a, b) = 0$ otherwise; $map(\cdot)$ is the mapping function corresponding to the best one-to-one assignment of clusters to labels implemented by the Hungarian algorithm \cite{kuhn1955hungarian}; Then NMI is computed by:
\begin{equation}
\label{eq:nmi}
	NMI = \frac{I(\mathbf{Y}; \mathbf{C})}{\frac{1}{2}(H(\mathbf{Y})+H(\mathbf{C}))}
\end{equation}
$I(\cdot; \cdot)$ and $H(\cdot)$ represent mutual information and entropy functionals, respectively. 

As for the classification task, we compute classification accuracy (ACC$_{cls}$) and F-score to report classification results, as shown below.

\begin{equation}
\label{eq:fscore}
Fscore = \frac{2 \times P \times R}{P + R}
\end{equation}
where $P = \frac{TP}{TP + FP}$; $TP$ and $FP$ are the number of true positives and the number of false positives, respectively; $R = \frac{TP}{TP+FN}$, where $FN$ is the number of false negatives. Higher values of all of the aforementioned metrics indicate better performance.

\section{Classification Results}

We evaluated the performance of all baseline models through classification tasks on the E-FMNIST and COIL-20 datasets, as summarized in Table \ref{tab:classification-result-append}. The results illuminate that, within the same experimental framework, the representations extracted by our method significantly enhance classification performance. Notably, in comparison to the second-best method, UNITER, our MRDD-$\bold{cs}$ approach demonstrated improvements of 4.59 and 4.58 in terms of Accuracy (ACC$_{cls}$) and F-score on the E-FMNIST dataset, respectively. These outcomes underscore that minimizing redundancy between view-consistent and view-specific representations proves advantageous in augmenting the effectiveness of downstream tasks.

\begin{figure*}[t]
\setlength{\abovecaptionskip}{0.2cm} 
\setlength{\belowcaptionskip}{-0.2cm}
    \centering
    \includegraphics[width=1\linewidth]{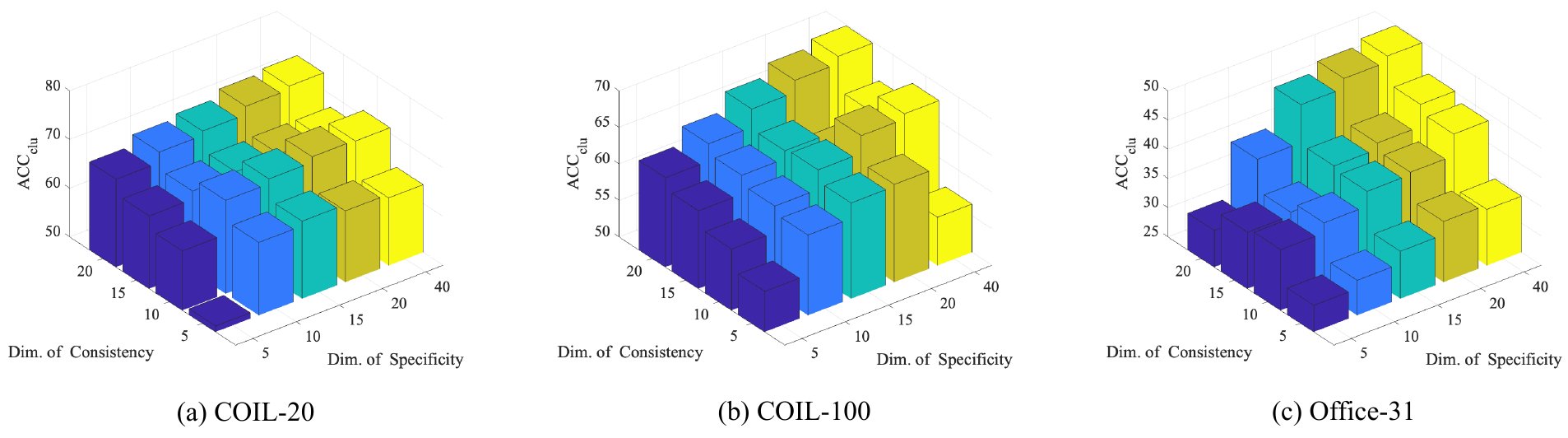}
   \caption{The clustering results (\%) of the different dimensions of consistency and specificity on the COIL-20, COIL-100, and Office-31 datasets. The x-axis represents the consistency dimension, the y-axis represents the specificity dimension, and the z-axis represents the clustering accuracy.}
   \label{fig:dimen}
\end{figure*}

\begin{table}[t]
\setlength{\abovecaptionskip}{0cm} 
\setlength{\belowcaptionskip}{-0.2cm} 
\setlength\tabcolsep{2pt}
\begin{center}
\scalebox{0.80}{
\begin{tabular}{lcccc}
\toprule
 & \multicolumn{2}{c}{E-FMNIST} & \multicolumn{2}{c}{COIL-20} \\
\cmidrule(lr){2-3} \cmidrule(lr){4-5}  
Method & ACC$_{cls}$ & F-Score & ACC$_{cls}$ & F-Score \\
\midrule
\textit{Random} & 9.99$\pm$0.13 & 9.99$\pm$0.13 & 4.60$\pm$0.67 & 3.11$\pm$0.46  \\
\rowcolor{Gray}Joint-VAE\cite{dupont2018learning} & 56.50$\pm$0.23 & 56.39$\pm$0.21 & 87.76 $\pm$2.00 & 84.24 $\pm$3.16 \\
$\beta$-VAE \cite{higgins2016beta} & 56.04$\pm$0.42 & 55.99$\pm$0.41 & 51.21$\pm$1.69 & 49.81$\pm$1.41 \\
\midrule
\rowcolor{Gray}CONAN$\dag$ \cite{ke2021conan} & 58.13$\pm$0.21 & 55.74$\pm$0.15 & 67.53$\pm$2.72 & 61.54$\pm$2.82 \\
CMC$\dag$ \cite{tian2020contrastive} &  67.43$\pm$0.13 & 64.85$\pm$0.17 & 89.16$\pm$0.01 & 89.15$\pm$0.01 \\
\rowcolor{Gray}Multi-VAE \cite{xu2021multi} & 81.54$\pm$0.38 & 79.43$\pm$0.24 & 90.39$\pm$1.12 & 89.32$\pm$1.53  \\
MIB~\cite{federici2020learning} &   75.33$\pm$0.05 & 73.80$\pm$0.05 & 59.72$\pm$2.29 & 53.99$\pm$2.03   \\
\rowcolor{Gray}DVIB \cite{bao2021disentangled} & 72.18$\pm$0.29 & 72.91$\pm$0.22 & 44.31$\pm$3.30 & 42.17$\pm$3.02 \\
UNITER \cite{xu2023untie} &  \underline{84.19$\pm$0.11} & \underline{84.10$\pm$0.11} & \underline{91.27$\pm$0.94} & \underline{90.58$\pm$1.01} \\
\midrule
\rowcolor{Gray}MRDD-$\bold{c}$ (Ours) &  82.51 $\pm$0.30 & 82.28$\pm$0.29 & 88.18$\pm$0.96 & 87.57$\pm$0.82  \\
MRDD-$\bold{cs}$ (Ours) & \textbf{88.78$\pm$0.22} & \textbf{88.68$\pm$0.18} & \textbf{95.97$\pm$0.56} & \textbf{96.15$\pm$0.88}  \\
\midrule
$\Delta$ SOTA  & \textcolor{commentcolor}{+4.59} & \textcolor{commentcolor}{+4.58} & \textcolor{commentcolor}{+4.7} & \textcolor{commentcolor}{5.57} \\

\bottomrule
\end{tabular}}
\end{center}
\caption{\textbf{Classification results (\%) on E-FMNIST and COIL-20 datasets.} \textbf{Bold} denotes the best results and \underline{underline} denotes the second-best. $\dag$ denotes we set the dimensionality of latent representations as 10. All results are reproduced using the official released code.}
\label{tab:classification-result-append}       
\end{table}

\begin{figure*}[t]
\setlength{\abovecaptionskip}{0.2cm} 
\setlength{\belowcaptionskip}{-0.2cm}
    \centering
    \includegraphics[width=1\linewidth]{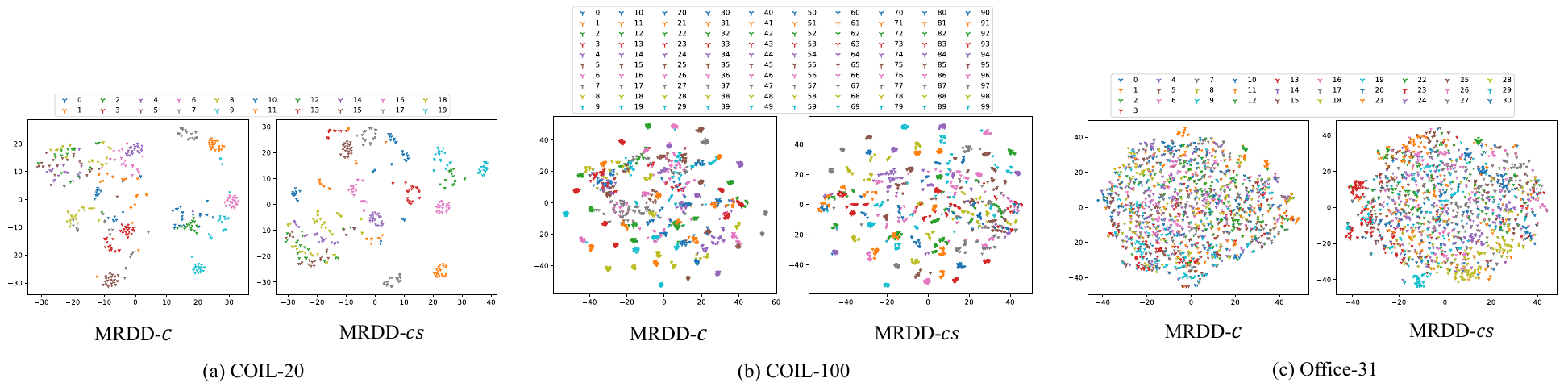}
   \caption{Visualization of the representations of MRDD-$\bold{c}$ and MRDD-$\bold{cs}$ using t-SNE~\cite{van2008visualizing} on the COIL-20, COIL-100, and Office-31.}
   \label{fig:tsne}
\end{figure*}

\begin{figure*}[t]
\setlength{\abovecaptionskip}{0.2cm} 
\setlength{\belowcaptionskip}{-0.2cm}
    \centering
    \includegraphics[width=1\linewidth]{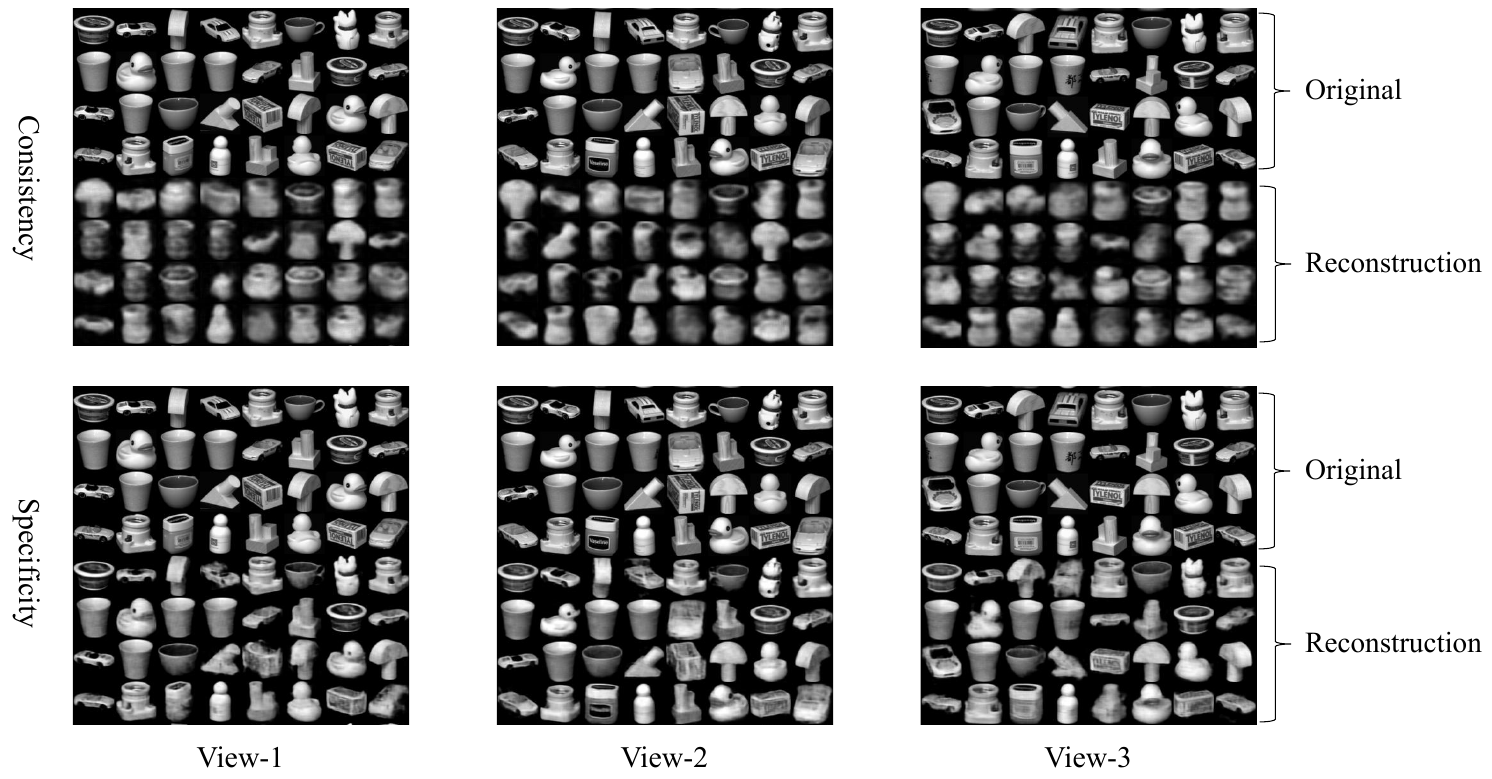}
   \caption{Visualization of reconstruction samples of consistency and specificity on the COIL-20 dataset.}
   \label{fig:recons-coil20}
\end{figure*}

\begin{figure*}[t]
\setlength{\abovecaptionskip}{0.2cm} 
\setlength{\belowcaptionskip}{-0.2cm}
    \centering
    \includegraphics[width=1\linewidth]{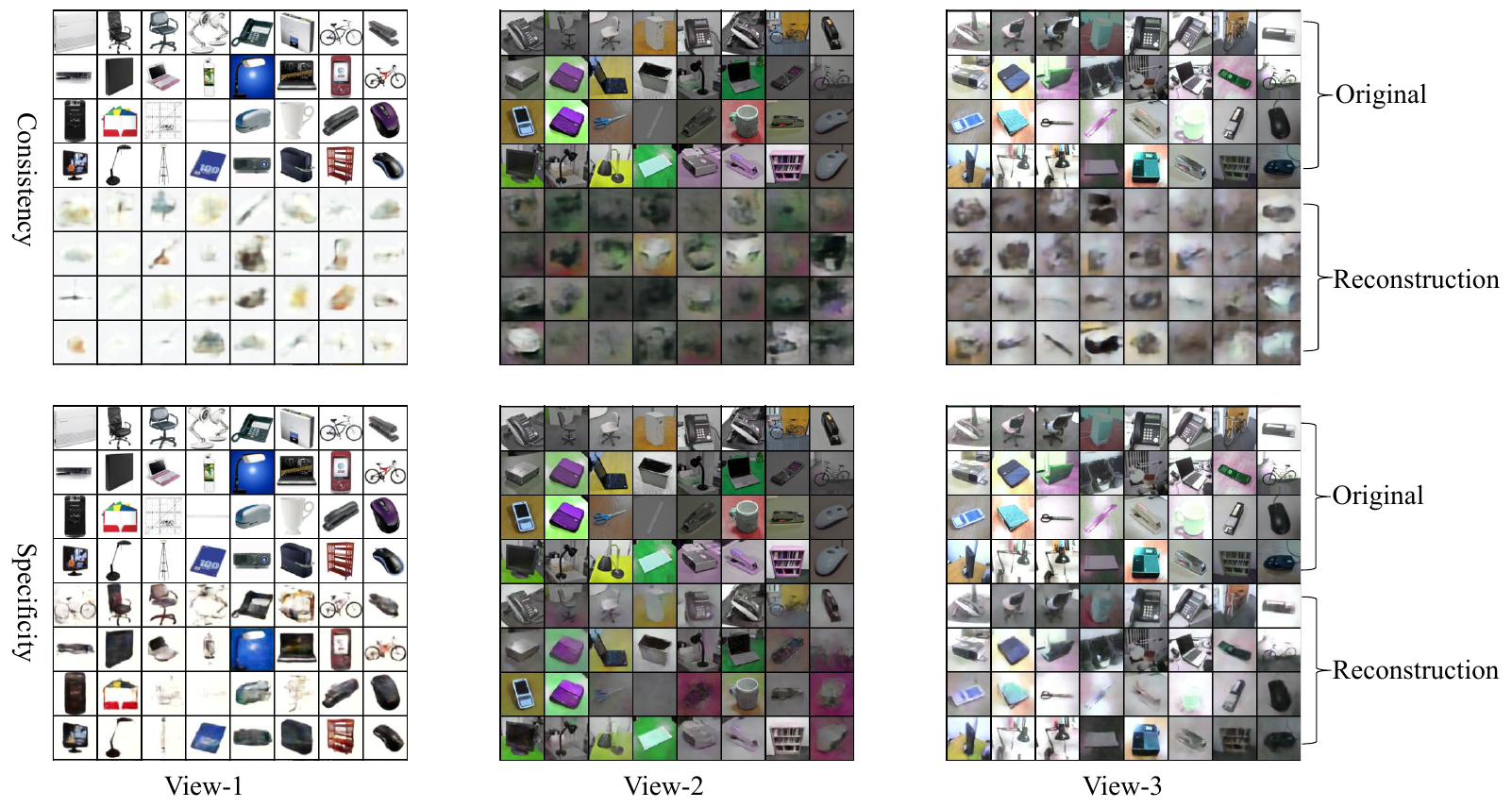}
   \caption{Visualization of reconstruction samples of consistency and specificity on the Office-31 dataset.}
   \label{fig:recons-office-31}
\end{figure*}

\section{Ablation Study}

\subsection{The dimension of consistency and specificity}

We investigate the impact of view-consistent and view-specific representations extracted by our method across various dimensions. The view-consistent representation dimensions are set within the range ${5, 10, 15, 20}$, while the view-specific representation dimension spans ${5, 10, 15, 20, 40}$. As illustrated in Fig.~\ref{fig:dimen}, the results show a positive correlation with the view-specific representation dimension when the view-consistent representation dimension is held constant. Specifically, when the dimensions of view-consistent representations are fixed at 20, a noticeable incremental relationship is observed between the dimensions of view-specific representations and clustering performance.

In contrast, when the dimensions of view-specific representations are fixed at 40, consistent representations do not exhibit a clear pattern of variation. We posit that the overall performance of our method is primarily influenced by the expressive capacity of view-consistent representations. Additionally, a marginal improvement in overall performance is noted when the dimensions of view-specific representations surpass those of view-consistent representations. This observation suggests a nuanced interplay between the dimensions of these representations and their impact on the performance of downstream tasks.

\section{Visualization}

We visualize the representations of MRDD-$\bold{c}$ and MRDD-$\bold{cs}$ on the COIL-20, COIL-100, and Office-31 dataset. Fig.~\ref{fig:tsne} indicates that view-consistent representations can distinguish different samples at a coarse level. However, after incorporating view-specific representations, the discriminative ability of the representations is enhanced, especially evident in the COIL-20 and COIL-100 dataset.

On the other hand, we demonstrate the reconstruction sampling of the COIL-20 and Office-31 datasets. As depicted in Fig.~\ref{fig:recons-coil20} and ~\ref{fig:recons-office-31}, reconstructing using only consistent representations results in the outline information of objects, indicating that the model has learned shared information among views. Furthermore, when incorporating view-specific representations, a significant improvement in reconstruction quality is observed. This suggests that view-specific representations contain information such as textures, details, and other nuanced aspects of objects.

{
    \small
    \bibliographystyle{ieeenat_fullname}
    \bibliography{main}

\begin{thebibliography}{66}
\providecommand{\natexlab}[1]{#1}
\providecommand{\url}[1]{\texttt{#1}}
\expandafter\ifx\csname urlstyle\endcsname\relax
  \providecommand{\doi}[1]{doi: #1}\else
  \providecommand{\doi}{doi: \begingroup \urlstyle{rm}\Url}\fi

\bibitem[Andrew et~al.(2013)Andrew, Arora, Bilmes, and Livescu]{andrew2013deep}
Galen Andrew, Raman Arora, Jeff Bilmes, and Karen Livescu.
\newblock Deep canonical correlation analysis.
\newblock In \emph{ICML}, pages 1247--1255, 2013.

\bibitem[Baevski et~al.(2020)Baevski, Zhou, Mohamed, and Auli]{baevski2020wav2vec}
Alexei Baevski, Yuhao Zhou, Abdelrahman Mohamed, and Michael Auli.
\newblock wav2vec 2.0: A framework for self-supervised learning of speech representations.
\newblock \emph{NeurIPS}, 33:\penalty0 12449--12460, 2020.

\bibitem[Bao(2021)]{bao2021disentangled}
Feng Bao.
\newblock Disentangled variational information bottleneck for multiview representation learning.
\newblock In \emph{CICAI}, pages 91--102, 2021.

\bibitem[Belghazi et~al.(2018)Belghazi, Baratin, Rajeshwar, Ozair, Bengio, Courville, and Hjelm]{belghazi2018mutual}
Mohamed~Ishmael Belghazi, Aristide Baratin, Sai Rajeshwar, Sherjil Ozair, Yoshua Bengio, Aaron Courville, and Devon Hjelm.
\newblock Mutual information neural estimation.
\newblock In \emph{ICML}, pages 531--540, 2018.

\bibitem[Bengio et~al.(2013)Bengio, Courville, and Vincent]{bengio2013representation}
Yoshua Bengio, Aaron Courville, and Pascal Vincent.
\newblock Representation learning: A review and new perspectives.
\newblock \emph{IEEE TPAMI}, 35\penalty0 (8):\penalty0 1798--1828, 2013.

\bibitem[Brbi{\'c} and Kopriva(2018)]{brbic2018multi}
Maria Brbi{\'c} and Ivica Kopriva.
\newblock Multi-view low-rank sparse subspace clustering.
\newblock \emph{Pattern Recognition}, 73:\penalty0 247--258, 2018.

\bibitem[Chao and Sun(2016)]{chao2016consensus}
Guoqing Chao and Shiliang Sun.
\newblock Consensus and complementarity based maximum entropy discrimination for multi-view classification.
\newblock \emph{Information Sciences}, 367:\penalty0 296--310, 2016.

\bibitem[Chen and Ho(2022)]{chen2022mm}
Jiawei Chen and Chiu~Man Ho.
\newblock Mm-vit: Multi-modal video transformer for compressed video action recognition.
\newblock In \emph{CVPR}, pages 1910--1921, 2022.

\bibitem[Cheng et~al.(2020)Cheng, Hao, Dai, Liu, Gan, and Carin]{cheng2020club}
Pengyu Cheng, Weituo Hao, Shuyang Dai, Jiachang Liu, Zhe Gan, and Lawrence Carin.
\newblock Club: A contrastive log-ratio upper bound of mutual information.
\newblock In \emph{ICML}, pages 1779--1788, 2020.

\bibitem[Dalal and Triggs(2005)]{dalal2005histograms}
Navneet Dalal and Bill Triggs.
\newblock Histograms of oriented gradients for human detection.
\newblock In \emph{CVPR}, pages 886--893, 2005.

\bibitem[Dupont(2018)]{dupont2018learning}
Emilien Dupont.
\newblock Learning disentangled joint continuous and discrete representations.
\newblock \emph{NeurIPS}, 31, 2018.

\bibitem[Federici et~al.(2020)Federici, Dutta, Forr{\'e}, Kushman, and Akata]{federici2020learning}
Marco Federici, Anjan Dutta, Patrick Forr{\'e}, Nate Kushman, and Zeynep Akata.
\newblock Learning robust representations via multi-view information bottleneck.
\newblock \emph{ICLR}, 2020.

\bibitem[Hangbo~Bao(2022)]{Bao0PW22}
Songhao Piao Furu~Wei Hangbo~Bao, Li~Dong.
\newblock Beit: {BERT} pre-training of image transformers.
\newblock In \emph{ICLR}, 2022.

\bibitem[He et~al.(2016)He, Zhang, Ren, and Sun]{he2016deep}
Kaiming He, Xiangyu Zhang, Shaoqing Ren, and Jian Sun.
\newblock Deep residual learning for image recognition.
\newblock In \emph{CVPR}, pages 770--778, 2016.

\bibitem[He et~al.(2022)He, Chen, Xie, Li, Doll{\'a}r, and Girshick]{he2022masked}
Kaiming He, Xinlei Chen, Saining Xie, Yanghao Li, Piotr Doll{\'a}r, and Ross Girshick.
\newblock Masked autoencoders are scalable vision learners.
\newblock In \emph{CVPR}, pages 16000--16009, 2022.

\bibitem[Higgins et~al.(2016)Higgins, Matthey, Pal, Burgess, Glorot, Botvinick, Mohamed, and Lerchner]{higgins2016beta}
Irina Higgins, Loic Matthey, Arka Pal, Christopher Burgess, Xavier Glorot, Matthew Botvinick, Shakir Mohamed, and Alexander Lerchner.
\newblock beta-vae: Learning basic visual concepts with a constrained variational framework.
\newblock In \emph{ICLR}, 2016.

\bibitem[Huang et~al.(2021)Huang, Zhou, Zhu, Zhang, Lv, and Peng]{huang2021deep}
Zhenyu Huang, Joey~Tianyi Zhou, Hongyuan Zhu, Changqing Zhang, Jiancheng Lv, and Xi Peng.
\newblock Deep spectral representation learning from multi-view data.
\newblock \emph{IEEE TIP}, 30:\penalty0 5352--5362, 2021.

\bibitem[Jin et~al.(2016)Jin, Chen, Chen, Xiong, and Hauptmann]{jin2016describing}
Qin Jin, Jia Chen, Shizhe Chen, Yifan Xiong, and Alexander Hauptmann.
\newblock Describing videos using multi-modal fusion.
\newblock In \emph{ACM MM}, pages 1087--1091, 2016.

\bibitem[Ke et~al.(2021)Ke, Hong, Zeng, Liu, Sun, and Xie]{ke2021conan}
Guanzhou Ke, Zhiyong Hong, Zhiqiang Zeng, Zeyi Liu, Yangjie Sun, and Yannan Xie.
\newblock Conan: contrastive fusion networks for multi-view clustering.
\newblock In \emph{IEEE International Conference on Big Data}, pages 653--660, 2021.

\bibitem[Ke et~al.(2022{\natexlab{a}})Ke, Hong, Yu, Zhang, and Liu]{ke2022efficient}
Guanzhou Ke, Zhiyong Hong, Wenhua Yu, Xin Zhang, and Zeyi Liu.
\newblock Efficient multi-view clustering networks.
\newblock \emph{Applied Intelligence}, 52\penalty0 (13):\penalty0 14918--14934, 2022{\natexlab{a}}.

\bibitem[Ke et~al.(2022{\natexlab{b}})Ke, Zhu, and Yu]{ke2022mori}
Guanzhou Ke, Yongqi Zhu, and Yang Yu.
\newblock Mori-ran: Multi-view robust representation learning via hybrid contrastive fusion.
\newblock In \emph{IEEE International Conference on Data Mining Workshops}, pages 467--474, 2022{\natexlab{b}}.

\bibitem[Ke et~al.(2023{\natexlab{a}})Ke, Chao, Wang, Xu, Zhu, and Yu]{ke2023clustering}
Guanzhou Ke, Guoqing Chao, Xiaoli Wang, Chenyang Xu, Yongqi Zhu, and Yang Yu.
\newblock A clustering-guided contrastive fusion for multi-view representation learning.
\newblock \emph{IEEE TCSVT}, 2023{\natexlab{a}}.

\bibitem[Ke et~al.(2023{\natexlab{b}})Ke, Yu, Chao, Wang, Xu, and He]{ke2023disentangling}
Guanzhou Ke, Yang Yu, Guoqing Chao, Xiaoli Wang, Chenyang Xu, and Shengfeng He.
\newblock Disentangling multi-view representations beyond inductive bias.
\newblock In \emph{ACM MM}, pages 2582--2590, 2023{\natexlab{b}}.

\bibitem[Kenton and Toutanova(2019)]{kenton2019bert}
Jacob Devlin Ming-Wei~Chang Kenton and Lee~Kristina Toutanova.
\newblock Bert: Pre-training of deep bidirectional transformers for language understanding.
\newblock In \emph{NAACL-HLT}, page~2, 2019.

\bibitem[Kingma and Welling(2014)]{KingmaW13}
Diederik~P. Kingma and Max Welling.
\newblock Auto-encoding variational bayes.
\newblock In \emph{ICLR}, 2014.

\bibitem[Kuhn(1955)]{kuhn1955hungarian}
Harold~W Kuhn.
\newblock The hungarian method for the assignment problem.
\newblock \emph{Naval research logistics quarterly}, 2\penalty0 (1-2):\penalty0 83--97, 1955.

\bibitem[Kumar et~al.(2011)Kumar, Rai, and Daume]{kumar2011co}
Abhishek Kumar, Piyush Rai, and Hal Daume.
\newblock Co-regularized multi-view spectral clustering.
\newblock \emph{NeurIPS}, 24:\penalty0 1413--1421, 2011.

\bibitem[Lan et~al.(2020)Lan, Chen, Goodman, Gimpel, Sharma, and Soricut]{LanCGGSS20}
Zhenzhong Lan, Mingda Chen, Sebastian Goodman, Kevin Gimpel, Piyush Sharma, and Radu Soricut.
\newblock {ALBERT:} {A} lite {BERT} for self-supervised learning of language representations.
\newblock In \emph{ICLR}, 2020.

\bibitem[Li et~al.(2019)Li, Yang, and Zhang]{LiYZ19}
Yingming Li, Ming Yang, and Zhongfei Zhang.
\newblock A survey of multi-view representation learning.
\newblock \emph{TKDE}, 31\penalty0 (10):\penalty0 1863--1883, 2019.

\bibitem[Li et~al.(2023)Li, Fan, Hu, Feichtenhofer, and He]{li2023scaling}
Yanghao Li, Haoqi Fan, Ronghang Hu, Christoph Feichtenhofer, and Kaiming He.
\newblock Scaling language-image pre-training via masking.
\newblock In \emph{CVPR}, pages 23390--23400, 2023.

\bibitem[Liu et~al.(2013)Liu, Wang, Gao, and Han]{liu2013multi}
Jialu Liu, Chi Wang, Jing Gao, and Jiawei Han.
\newblock Multi-view clustering via joint nonnegative matrix factorization.
\newblock In \emph{ICDM}, pages 252--260, 2013.

\bibitem[Liu and Tuzel(2016)]{liu2016coupled}
Ming-Yu Liu and Oncel Tuzel.
\newblock Coupled generative adversarial networks.
\newblock \emph{NeurIPS}, 29, 2016.

\bibitem[Liu et~al.(2019)Liu, Ott, Goyal, Du, Joshi, Chen, Levy, Lewis, Zettlemoyer, and Stoyanov]{liu2019roberta}
Yinhan Liu, Myle Ott, Naman Goyal, Jingfei Du, Mandar Joshi, Danqi Chen, Omer Levy, Mike Lewis, Luke Zettlemoyer, and Veselin Stoyanov.
\newblock Roberta: A robustly optimized bert pretraining approach.
\newblock \emph{arXiv preprint arXiv:1907.11692}, 2019.

\bibitem[Lowe(1999)]{lowe1999object}
David~G Lowe.
\newblock Object recognition from local scale-invariant features.
\newblock In \emph{ICCV}, pages 1150--1157, 1999.

\bibitem[Nene et~al.(1996)Nene, Nayar, Murase, et~al.]{nene1996columbia}
Sameer~A Nene, Shree~K Nayar, Hiroshi Murase, et~al.
\newblock Columbia object image library (coil-20).
\newblock 1996.

\bibitem[Paszke et~al.(2019)Paszke, Gross, Massa, Lerer, Bradbury, Chanan, Killeen, Lin, Gimelshein, Antiga, et~al.]{paszke2019pytorch}
Adam Paszke, Sam Gross, Francisco Massa, Adam Lerer, James Bradbury, Gregory Chanan, Trevor Killeen, Zeming Lin, Natalia Gimelshein, Luca Antiga, et~al.
\newblock Pytorch: An imperative style, high-performance deep learning library.
\newblock \emph{NeurIPS}, 32:\penalty0 8026--8037, 2019.

\bibitem[Rasiwasia et~al.(2010)Rasiwasia, Costa~Pereira, Coviello, Doyle, Lanckriet, Levy, and Vasconcelos]{rasiwasia2010new}
Nikhil Rasiwasia, Jose Costa~Pereira, Emanuele Coviello, Gabriel Doyle, Gert~RG Lanckriet, Roger Levy, and Nuno Vasconcelos.
\newblock A new approach to cross-modal multimedia retrieval.
\newblock In \emph{ACM MM}, pages 251--260, 2010.

\bibitem[Ren et~al.(2021)Ren, Du, Lv, Han, and He]{ren2021learning}
Sucheng Ren, Yong Du, Jianming Lv, Guoqiang Han, and Shengfeng He.
\newblock Learning from the master: Distilling cross-modal advanced knowledge for lip reading.
\newblock In \emph{CVPR}, pages 13325--13333, 2021.

\bibitem[Ren et~al.(2022)Ren, Gao, Hua, Xue, Tian, He, and Zhao]{ren2022co}
Sucheng Ren, Zhengqi Gao, Tianyu Hua, Zihui Xue, Yonglong Tian, Shengfeng He, and Hang Zhao.
\newblock Co-advise: Cross inductive bias distillation.
\newblock In \emph{CVPR}, pages 16773--16782, 2022.

\bibitem[Saenko et~al.(2010)Saenko, Kulis, Fritz, and Darrell]{saenko2010adapting}
Kate Saenko, Brian Kulis, Mario Fritz, and Trevor Darrell.
\newblock Adapting visual category models to new domains.
\newblock In \emph{ECCV}, pages 213--226, 2010.

\bibitem[Sun and Chao(2013)]{sun2013multi}
Shiliang Sun and Guoqing Chao.
\newblock Multi-view maximum entropy discrimination.
\newblock In \emph{IJCAI}, pages 1706--1712, 2013.

\bibitem[Sun et~al.(2019)Sun, Cheng, Min, and Jing]{sun2019self}
Xiukun Sun, Miaomiao Cheng, Chen Min, and Liping Jing.
\newblock Self-supervised deep multi-view subspace clustering.
\newblock In \emph{ACML}, pages 1001--1016, 2019.

\bibitem[Tian et~al.(2020)Tian, Krishnan, and Isola]{tian2020contrastive}
Yonglong Tian, Dilip Krishnan, and Phillip Isola.
\newblock Contrastive multiview coding.
\newblock In \emph{ECCV}, pages 776--794, 2020.

\bibitem[Trosten et~al.(2021)Trosten, Lokse, Jenssen, and Kampffmeyer]{trosten2021reconsidering}
Daniel~J Trosten, Sigurd Lokse, Robert Jenssen, and Michael Kampffmeyer.
\newblock Reconsidering representation alignment for multi-view clustering.
\newblock In \emph{CVPR}, pages 1255--1265, 2021.

\bibitem[Van~der Maaten and Hinton(2008)]{van2008visualizing}
Laurens Van~der Maaten and Geoffrey Hinton.
\newblock Visualizing data using t-sne.
\newblock \emph{Journal of machine learning research}, 9\penalty0 (11), 2008.

\bibitem[Vaswani et~al.(2017)Vaswani, Shazeer, Parmar, Uszkoreit, Jones, Gomez, Kaiser, and Polosukhin]{vaswani2017attention}
Ashish Vaswani, Noam Shazeer, Niki Parmar, Jakob Uszkoreit, Llion Jones, Aidan~N Gomez, {\L}ukasz Kaiser, and Illia Polosukhin.
\newblock Attention is all you need.
\newblock \emph{NeurIPS}, 30, 2017.

\bibitem[Vincent et~al.(2008)Vincent, Larochelle, Bengio, and Manzagol]{vincent2008extracting}
Pascal Vincent, Hugo Larochelle, Yoshua Bengio, and Pierre-Antoine Manzagol.
\newblock Extracting and composing robust features with denoising autoencoders.
\newblock In \emph{ICML}, pages 1096--1103, 2008.

\bibitem[Wang et~al.(2015)Wang, Arora, Livescu, and Bilmes]{wang2015deep}
Weiran Wang, Raman Arora, Karen Livescu, and Jeff Bilmes.
\newblock On deep multi-view representation learning.
\newblock In \emph{ICML}, pages 1083--1092, 2015.

\bibitem[Wang et~al.(2019)Wang, Lei, Guo, Zhang, Shi, and Li]{wang2019multi}
Xiaobo Wang, Zhen Lei, Xiaojie Guo, Changqing Zhang, Hailin Shi, and Stan~Z Li.
\newblock Multi-view subspace clustering with intactness-aware similarity.
\newblock \emph{Pattern Recognition}, 88:\penalty0 50--63, 2019.

\bibitem[Wang et~al.(2021)Wang, Zhang, Shen, Kong, and Li]{wang2021dense}
Xinlong Wang, Rufeng Zhang, Chunhua Shen, Tao Kong, and Lei Li.
\newblock Dense contrastive learning for self-supervised visual pre-training.
\newblock In \emph{CVPR}, pages 3024--3033, 2021.

\bibitem[Wang et~al.(2022)Wang, Chen, Tang, Wu, and Zhu]{wang2022disentangled}
Xin Wang, Hong Chen, Si'ao Tang, Zihao Wu, and Wenwu Zhu.
\newblock Disentangled representation learning.
\newblock \emph{arXiv preprint arXiv:2211.11695}, 2022.

\bibitem[Wang et~al.(2024)Wang, Wang, Ke, Wang, and Hong]{wang2024knowledge}
Xiaoli Wang, Yongli Wang, Guanzhou Ke, Yupeng Wang, and Xiaobin Hong.
\newblock Knowledge distillation-driven semi-supervised multi-view classification.
\newblock \emph{Information Fusion}, 103:\penalty0 102098, 2024.

\bibitem[Wang et~al.(2018)Wang, Wu, Lin, and Gao]{wang2018multiview}
Yang Wang, Lin Wu, Xuemin Lin, and Junbin Gao.
\newblock Multiview spectral clustering via structured low-rank matrix factorization.
\newblock \emph{IEEE TNNLS}, 29\penalty0 (10):\penalty0 4833--4843, 2018.

\bibitem[Wei et~al.(2022)Wei, Fan, Xie, Wu, Yuille, and Feichtenhofer]{wei2022masked}
Chen Wei, Haoqi Fan, Saining Xie, Chao-Yuan Wu, Alan Yuille, and Christoph Feichtenhofer.
\newblock Masked feature prediction for self-supervised visual pre-training.
\newblock In \emph{CVPR}, pages 14668--14678, 2022.

\bibitem[Xiao et~al.(2017)Xiao, Rasul, and Vollgraf]{xiao2017fashion}
Han Xiao, Kashif Rasul, and Roland Vollgraf.
\newblock Fashion-mnist: a novel image dataset for benchmarking machine learning algorithms.
\newblock \emph{arXiv preprint arXiv:1708.07747}, 2017.

\bibitem[Xie et~al.(2023)Xie, Zhang, Xu, Zhu, and He]{xie2023towards}
Yi Xie, Huaidong Zhang, Xuemiao Xu, Jianqing Zhu, and Shengfeng He.
\newblock Towards a smaller student: Capacity dynamic distillation for efficient image retrieval.
\newblock In \emph{CVPR}, pages 16006--16015, 2023.

\bibitem[Xie et~al.(2024)Xie, Lin, Cai, Xu, Zhang, Du, and He]{xie24cvpr}
Yi Xie, Yihong Lin, Wenjie Cai, Xuemiao Xu, Huaidong Zhang, Yong Du, and Shengfeng He.
\newblock D3still: Decoupled differential distillation for asymmetric image retrieval.
\newblock In \emph{CVPR}, 2024.

\bibitem[Xu et~al.(2023{\natexlab{a}})Xu, Zhao, Zhao, Guan, Yang, Chen, and Song]{xu2023progressive}
Cai Xu, Wei Zhao, Jinglong Zhao, Ziyu Guan, Yaming Yang, Long Chen, and Xiangyu Song.
\newblock Progressive deep multi-view comprehensive representation learning.
\newblock In \emph{AAAI}, pages 10557--10565, 2023{\natexlab{a}}.

\bibitem[Xu et~al.(2021)Xu, Ren, Tang, Pu, Zhu, Zeng, and He]{xu2021multi}
Jie Xu, Yazhou Ren, Huayi Tang, Xiaorong Pu, Xiaofeng Zhu, Ming Zeng, and Lifang He.
\newblock Multi-vae: Learning disentangled view-common and view-peculiar visual representations for multi-view clustering.
\newblock In \emph{CVPR}, pages 9234--9243, 2021.

\bibitem[Xu et~al.(2022)Xu, Tang, Ren, Peng, Zhu, and He]{xu2022multi}
Jie Xu, Huayi Tang, Yazhou Ren, Liang Peng, Xiaofeng Zhu, and Lifang He.
\newblock Multi-level feature learning for contrastive multi-view clustering.
\newblock In \emph{CVPR}, pages 16051--16060, 2022.

\bibitem[Xu et~al.(2023{\natexlab{b}})Xu, Ren, Shi, Shen, and Zhu]{xu2023untie}
Jie Xu, Yazhou Ren, Xiaoshuang Shi, Heng~Tao Shen, and Xiaofeng Zhu.
\newblock Untie: Clustering analysis with disentanglement in multi-view information fusion.
\newblock \emph{Information Fusion}, 100:\penalty0 101937, 2023{\natexlab{b}}.

\bibitem[Yan et~al.(2023)Yan, Zhang, Lv, Tang, Yue, Liao, and Lin]{yan2023gcfagg}
Weiqing Yan, Yuanyang Zhang, Chenlei Lv, Chang Tang, Guanghui Yue, Liang Liao, and Weisi Lin.
\newblock Gcfagg: Global and cross-view feature aggregation for multi-view clustering.
\newblock In \emph{CVPR}, pages 19863--19872, 2023.

\bibitem[Zhang et~al.(2019)Zhang, Liu, and Fu]{zhang2019ae2}
Changqing Zhang, Yeqing Liu, and Huazhu Fu.
\newblock Ae2-nets: Autoencoder in autoencoder networks.
\newblock In \emph{CVPR}, pages 2577--2585, 2019.

\bibitem[Zhao et~al.(2017)Zhao, Ding, and Fu]{zhao2017multi}
Handong Zhao, Zhengming Ding, and Yun Fu.
\newblock Multi-view clustering via deep matrix factorization.
\newblock In \emph{AAAI}, 2017.

\bibitem[Zhou and Shen(2020)]{zhou2020end}
Runwu Zhou and Yi-Dong Shen.
\newblock End-to-end adversarial-attention network for multi-modal clustering.
\newblock In \emph{CVPR}, pages 14619--14628, 2020.

\bibitem[Zhou et~al.(2020)Zhou, Sun, Zhang, Anguelov, Gao, Ouyang, Guo, Ngiam, and Vasudevan]{zhou2020end-3d}
Yin Zhou, Pei Sun, Yu Zhang, Dragomir Anguelov, Jiyang Gao, Tom Ouyang, James Guo, Jiquan Ngiam, and Vijay Vasudevan.
\newblock End-to-end multi-view fusion for 3d object detection in lidar point clouds.
\newblock In \emph{CoRL}, pages 923--932, 2020.

\end{thebibliography}
}


\end{document}